\documentclass{article}
\usepackage{iclr2026_conference,times}

\usepackage{url}
\usepackage{caption}

\usepackage{graphicx}
\usepackage{amsmath}
\usepackage{amssymb}
\usepackage{fontawesome5} %
\usepackage{booktabs}
\usepackage{latexsym}
\usepackage[T1]{fontenc}
\usepackage{float}
\usepackage[utf8]{inputenc}
\usepackage{microtype}
\usepackage{inconsolata}
\usepackage{url}
\usepackage{marvosym}
\usepackage{xcolor}
\usepackage{colortbl}
\usepackage{booktabs}
\usepackage[most]{tcolorbox}

\usepackage{tabularx}
\usepackage{booktabs}
\usepackage{enumitem}
\usepackage{xltabular}

\usepackage{longtable}
\usepackage{booktabs}
\usepackage{enumitem}

\usepackage{supertabular}
\usepackage{booktabs}
\usepackage{enumitem}
\usepackage{caption}

\definecolor{emerald}{rgb}{0.31, 0.78, 0.47}
\usepackage{tabularx}
\usepackage{booktabs}
\usepackage{enumitem}
\usepackage{xltabular}

\usepackage{longtable}
\usepackage{booktabs}
\usepackage{enumitem}

\usepackage{supertabular}
\usepackage{booktabs}
\usepackage{enumitem}
\usepackage{caption}

\usepackage{graphicx}
\usepackage{titletoc}
\usepackage{subcaption}  
\usepackage{dsfont}
\usepackage{wrapfig}
\usepackage{overpic}
\usepackage{cuted} 
\usepackage[export]{adjustbox}
\usepackage{verbatim}
\usepackage{amsmath} 
\usepackage{multirow} 
\usepackage[title]{appendix}
\usepackage{pifont}
\usepackage{subcaption}
\usepackage{xspace}
\usepackage{array}
\usepackage{enumitem}

\definecolor{softred}{RGB}{252,240,234}
\definecolor{softgreen}{RGB}{228,246,252}
\usepackage{colortbl,xcolor}
\usepackage[most]{tcolorbox}

\usepackage[most]{tcolorbox}
\tcbuselibrary{breakable, skins, listings}
\usepackage{caption} %

\tcbset{
  colback=gray!5,
  colframe=gray!30,
  boxrule=0.3pt,
  arc=2pt,
  left=4pt,
  right=4pt,
  top=2pt,
  bottom=2pt
}

\usepackage{booktabs}
\usepackage{graphicx}
\usepackage[most]{tcolorbox}
\usepackage{array}
\usepackage{enumitem}

\tcbset{
  boxrule=0.4pt,
  arc=2pt,
  colframe=black,
  colback=white
}

\newcommand{\datasrcicon}[1]{%
  \adjustbox{valign=m}{\includegraphics[height=1.6em]{#1}}%
  \xspace}

\newcommand{\humanlogo}{\datasrcicon{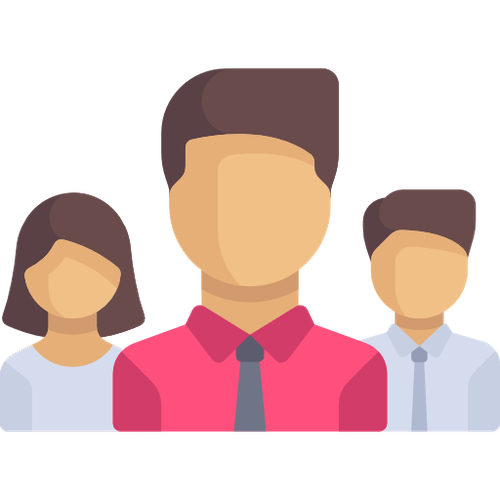}}
\newcommand{\chatgptlogo}{\datasrcicon{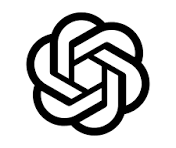}}

\definecolor{cvprblue}{rgb}{0.21,0.49,0.74}
\definecolor{cvprblue}{RGB}{0,90,150}

\definecolor{metablue}{HTML}{0064E0}

\newcommand{\rparagraph}[1]{\vspace{1.2mm}\noindent\textbf{#1}}

\renewcommand{\cite}{\citep}

\title{%
\makebox[\textwidth][l]{%
  \includegraphics[height=0.5cm]{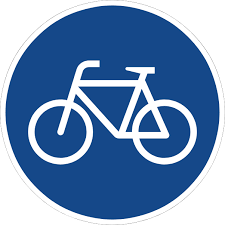}\hspace{0.5em}%
  From Steering to Pedalling: Do Autonomous%
}\\
\makebox[\textwidth][l]{ Driving VLMs Generalize to Cyclist-}\\
\makebox[\textwidth][l]{Assistive Spatial Perception and Planning?}
}

\usepackage[pagebackref,breaklinks,colorlinks,allcolors=metablue,urlcolor=metablue]{hyperref}

\author{
    Krishna Kanth Nakka\thanks{Core Contributor} 
    \quad {and} \quad
    Vedasri Nakka\thanks{Data Curation and Annotation Support}
    \vspace{0.2cm} \\
    Munich, Bavaria, Germany \\
    {\tt krishkanth.92@gmail.com}\\
    {~\url{https://krishnakanthnakka.github.io/CyclingVQA}}
}

\iclrfinalcopy
\begin{document}

\maketitle

\begin{abstract}
  Cyclists often encounter safety-critical situations in urban traffic, highlighting the need for assistive systems that support safe and informed decision-making. Recently, vision--language models (VLMs) have demonstrated strong performance on autonomous driving benchmarks, suggesting their potential for general traffic understanding and navigation-related reasoning. However, existing evaluations are predominantly vehicle-centric and fail to assess perception and reasoning from a cyclist-centric viewpoint. To address this gap, we introduce \textit{CyclingVQA}, a diagnostic benchmark designed to probe perception, spatio-temporal understanding, and traffic-rule-to-lane reasoning from a cyclist’s perspective. Evaluating \textbf{31+} recent VLMs spanning general-purpose, spatially enhanced, and autonomous-driving-specialized models, we find that current models demonstrate encouraging capabilities, while also revealing clear areas for improvement in cyclist-centric perception and reasoning, particularly in interpreting cyclist-specific traffic cues and associating signs with the correct navigational lanes. Notably, several driving-specialized models underperform strong generalist VLMs, indicating limited transfer from vehicle-centric training to cyclist-assistive scenarios. Finally, through systematic error analysis, we identify recurring failure modes to guide the development of more effective cyclist-assistive intelligent systems.
\end{abstract}

\section{Introduction}

Cycling has emerged as a cornerstone of sustainable urban mobility, offering a low-carbon and health-conscious alternative to motorized transport. This importance is reflected in its widespread adoption across Europe, where cycling accounts for a substantial fraction of daily commutes, including approximately $41\%$ in the Netherlands and $15\%$ in Germany~\cite{euronews_cycling_europe_2023}. Moreover, because cycling does not require formal licensing, it remains accessible to a broad demographic with varying levels of traffic experience. Yet this same accessibility, however, comes at a cost: cyclists are among the most vulnerable road users. In Germany alone, traffic accidents in 2023 resulted in several hundred cyclist fatalities and tens of thousands of serious injuries~\cite{ziv_single_bicycle_accidents}. Despite these risks, a significant technological disparity persists. While autonomous driving (AD) research~\cite{zhou2024vision,xie2025vlms,cui2024survey,yang2023llm4drive,jiang2025alphadrive} has made substantial progress in vehicle-centric perception and decision-making, cyclists still lack dedicated, perception-driven assistive systems to support navigation in complex or high-risk traffic scenarios. For example, at busy intersections, cyclists may struggle with interpreting traffic signs, selecting the correct lane, and making safe crossing decisions, underscoring the value of cyclist-assistive systems with strong perceptual and reasoning capabilities.

Intutively, cyclist-assistive intelligence requires many of the same perceptual and spatial reasoning capabilities as autonomous driving, including recognizing traffic signs, identifying lane boundaries, and reasoning over road layout. Recently, vision--language models (VLMs)~\cite{team2025gemma, li2024llava, liu2024llavanext, bai2025qwen2} have demonstrated strong capabilities across a wide range of multimodal tasks. Alongside this progress, a substantial body of work has assessed and improved spatial reasoning in VLMs through dedicated benchmarks~\cite{fu2024blink, liu2024ocrbench, yue2024mmmu, lu2023mathvista, kazemzadeh2014referitgame, yang2025visual, batra2025spatialthinker, ogezi2025spare, gan2025foundationmotion, shen2025fine}. Separately, VLMs have been adapted to autonomous driving scenarios~\cite{li2025recogdrive, ma2023dolphins, drivemm, azzolini2025cosmos, tian2024drivevlm, ishaq2025drivelmm}, enabling vehicle-centric perception, reasoning, and planning tasks~\cite{corbiere2025retrieval, ishaq2025drivelmm}.
However, it remains unclear whether these capabilities of AD-VLMs transfer to cyclist-assistive settings, where perspective, infrastructure, and traffic rules differ. This motivates a central research question: \emph{To what extent do existing VLMs, particularly those optimized for autonomous driving, generalize to the spatial perception and traffic understanding demands of cyclist-perspective scenarios?}

\begin{figure}[!t]
    \centering
    \begin{minipage}{0.4\linewidth}
        \centering
        \includegraphics[width=\linewidth]{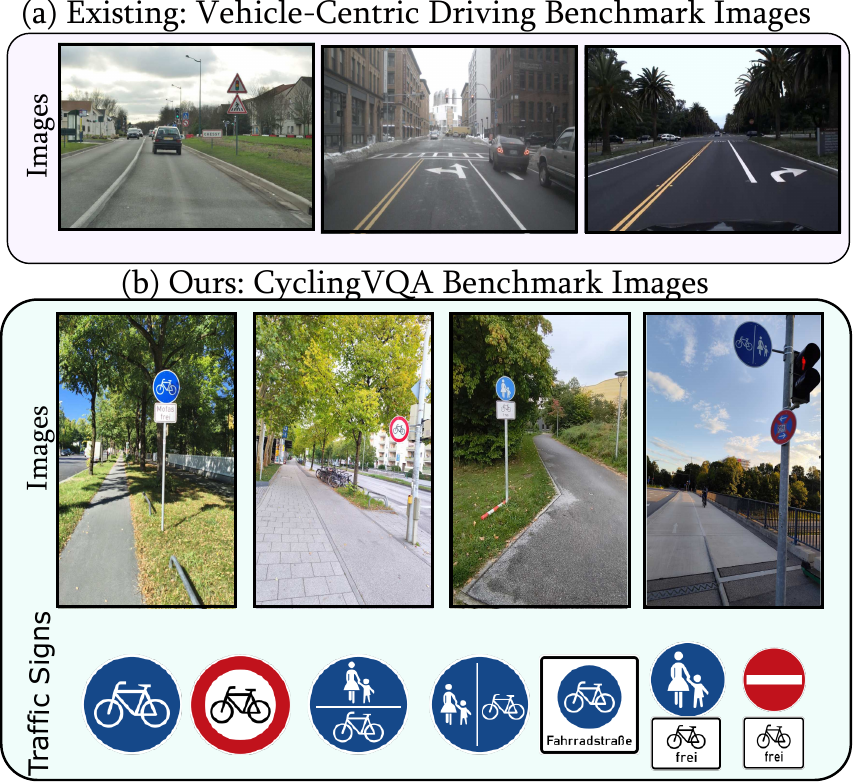}
    \end{minipage}
    \hfill
    \begin{minipage}{0.55\linewidth}
        \caption{Comparison between vehicle-centric driving benchmarks~\cite{corbiere2025retrieval,tian2025nuscenes,li2025fine}, which predominantly focus on road-level perspectives, and our cyclist-centric viewpoint, highlighting differences in camera perspective and the presence of cycling-specific traffic signage. See Appendix~\ref{sec:qualitativeresults} for further examples from our dataset.}
        \label{fig:teaser}
    \end{minipage}
\end{figure}

Unsurprisingly, existing traffic benchmarks for VLMs~\cite{cao2024maplm, ishaq2025drivelmm, wei2025driveqa, ghosh2025roadwork, tian2025nuscenes, li2025fine} remain predominantly vehicle-centric, focusing on motorized traffic flows from a driver’s perspective. While these benchmarks are highly relevant,  a cyclist’s viewpoint,  as illustrated in Figure~\ref{fig:teaser}, differs in several important respects from that of a car, introducing distinct navigational constraints such as bicycle-only lanes, cyclist-specific signage, and the negotiation of shared pedestrian spaces. Consequently, images derived from vehicle-centric datasets render such data ill-suited for evaluating whether VLMs can support cyclist-assistive perception and decision-making, particularly in scenarios that require fine-grained associations between traffic signs, lanes, and permitted actions.

To bridge this gap, we introduce \textnormal{CyclingVQA}, a first effort toward evaluating cyclist-perspective spatial perception and navigation-relevant reasoning in urban traffic scenes. CyclingVQA provides a granular assessment from a cyclist’s egocentric perspective, serving as a foundational step toward cyclist-assistive intelligent systems. The benchmark comprises \textbf{2,009} multiple-choice visual question--answer pairs derived from \textbf{695} images captured in real-world urban cycling environments in Munich. As a central contribution of this work, we evaluate a diverse suite of state-of-the-art VLMs—including general-purpose models, spatially enhanced architectures~\cite{yang2025visual, batra2025spatialthinker, cho2025perceptionlm}, and autonomous-driving-specialized models~\cite{li2025recogdrive, azzolini2025cosmos}. While several models exhibit promising zero-shot performance, they still fall short in challneging tasks. In particular, multiple driving-specialized VLMs struggle to interpret cyclist-specific traffic cues and to associate signs with the correct navigational lanes, often underperforming strong generalist baselines. We believe these empirical findings, supported by a systematic failure mode analysis, highlight important limitations in current traffic-oriented VLMs and can help guide the development of more effective cyclist-assistive systems. In summary, our contributions are as follows:

\begin{enumerate}
  \item \textbf{Cyclist-Centric Benchmark:} We introduce CyclingVQA, a cyclist-centric benchmark comprising 2,009 multiple-choice visual question--answer pairs derived from 695 real-world egocentric images (\S~\ref{sec:qageneration}).
    \item \textbf{Specialized Spatial Evaluation:} We define eight evaluation tasks that probe cyclist-centric spatial perception, traffic rule compliance, and navigation-relevant reasoning in complex urban environments (\S~\ref{sec:tasks}).
    \item \textbf{Comprehensive VLM Evaluation:} We benchmark state-of-the-art VLMs, including general-purpose, spatially enhanced~\cite{yang2025visual,batra2025spatialthinker,cho2025perceptionlm}, and autonomous-driving-focused models~\cite{li2025recogdrive,azzolini2025cosmos,cosmosreason2,ishaq2025drivelmm}, revealing substantial room for improvement in cyclist-centric reasoning (\S~\ref{sec:results}).
    \item \textbf{Systematic Failure Analysis:} We conduct a comprehensive analysis of recurring failure modes, providing insights and guidance for future cyclist-assistive intelligent systems (\S~\ref{sec:erroranalysis}).
\end{enumerate}

\section{Related Work}\label{related_work}

We provide a brief review of related benchmarks and specialist VLMs in the context of spatial intelligencedd and autonomous driving.

\rparagraph{Spatial Reasoning and Driving Benchmarks.}
The rapid emergence of benchmarks over the past two years has significantly advanced the evaluation of multimodal spatial understanding in VLMs. General-purpose spatial reasoning benchmarks~\cite{fu2024blink, cheng2024spatialrgpt, wang2024picture, jia2025omnispatial, ma20253dsrbench, kamath2023s, stogiannidis2025mind} typically assess a model’s understanding of object relationships, geometry, orientation, and relative positioning through visual question answering tasks. While foundational, these benchmarks, albeit effective for general spatial reasoning, predominantly rely on generic imagery, leaving the specific perceptual and semantic challenges of cyclist-assistive traffic scenarios largely unexplored. Our benchmark extend the above spatial reasoning benchmarks by introducing a dedicated suite of tasks tailored to cyclist-centric intelligent assistance.

Expanding on this, a parallel line of research has introduced driving benchmarks such as DrivingVQA~\cite{corbiere2025retrieval} and DriveQA~\cite{wei2025driveqa}, which evaluate models from a driver’s perspective on tasks ranging from right-of-way reasoning to decision-making. Recent efforts further emphasize specialized settings, including construction-zone understanding~\cite{ghosh2025roadwork}, fine-grained perception~\cite{li2025fine}, and safety-critical robustness~\cite{xing2024autotrust}. Most of these benchmarks are derived from large-scale, vehicle-centric datasets such as Waymo~\cite{sun2020scalability} and nuScenes~\cite{caesar2020nuscenes}. However, as we mentioned, these datasets are captured exclusively from motor vehicles, they do not capture key spatial semantics of the cycling environment, including dedicated bicycle lanes, shared-use paths, and cyclist-specific traffic signals. As a result, existing benchmarks do not adequately represent the egocentric viewpoints and navigational constraints inherent to cycling, limiting their suitability for evaluating cyclist-assistive reasoning. Our work addresses this gap by introducing a benchmark explicitly designed from the cyclist’s perspective.

\begin{figure*}[t!]
    \centering
    \includegraphics[width=0.99\textwidth]{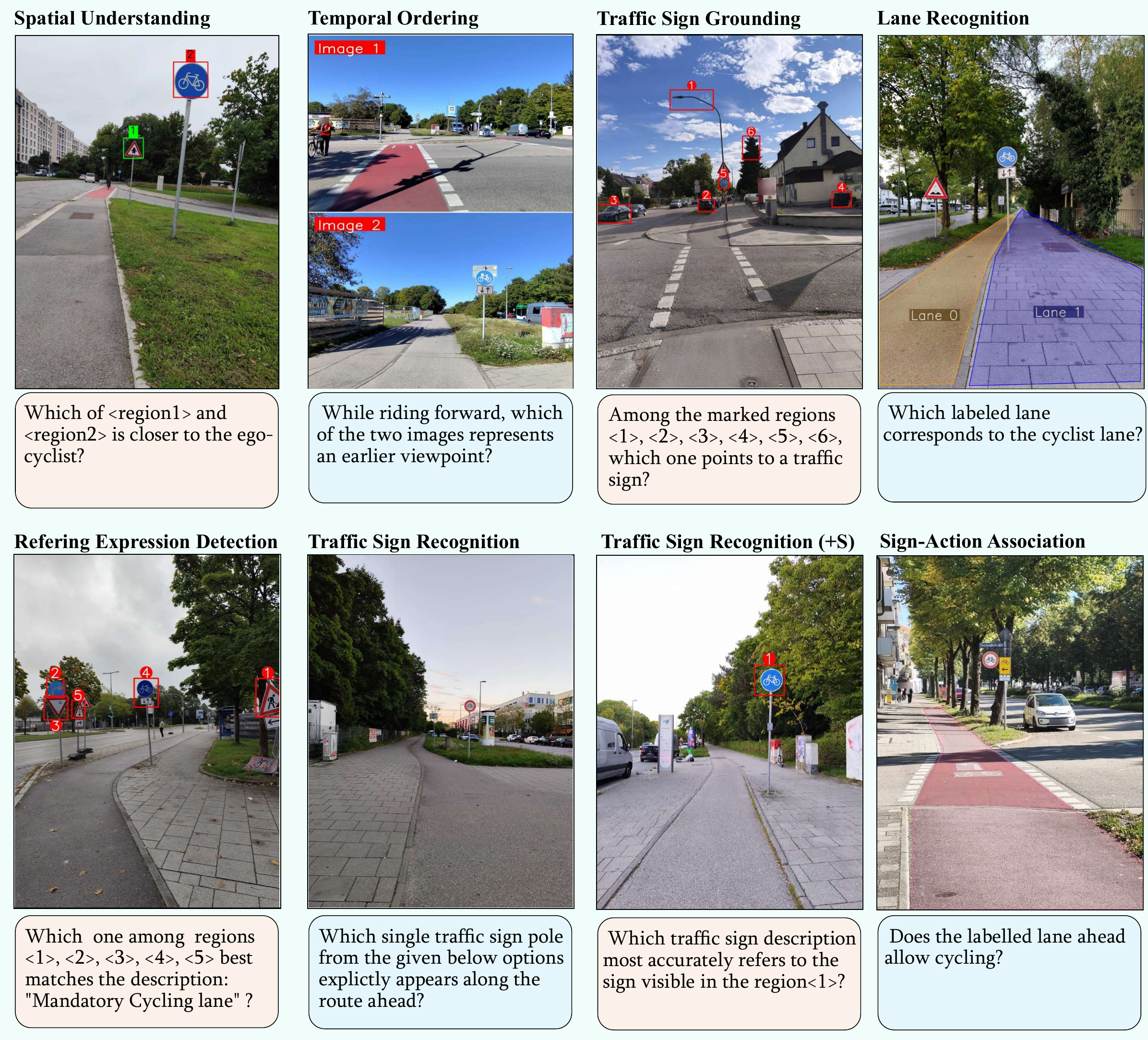}

    \vspace{-0.2cm}
\caption{\textbf{Benchmark tasks.} Illustration of the eight benchmark tasks in {CyclingVQA}, showing example question prompts together with visual inputs augmented by lane annotations and bounding-box supervision.}
    \label{fig:tasks_illustration}
    \vspace{-0.3cm}
\end{figure*}

\noindent\textbf{Specialist VLMs.}
The development of benchmarks has been accompanied by the emergence of specialist VLMs with enhanced spatial awareness. Spatially enhanced models~\cite{yang2025visual, batra2025spatialthinker, cho2025perceptionlm} are often fine-tuned on grounded datasets such as SpatialVQA~\cite{chen2024spatialvlm} to improve spatial grounding and relative distance understanding. Besides, autonomous-driving-focused VLMs~\cite{drivemm, li2025recogdrive, ma2023dolphins, azzolini2025cosmos} are optimized for traffic scene perception and planning in vehicle-centric settings. While these specialist models achieve strong performance within their target domains, their ability to generalize to the distinct viewpoints, constraints, and traffic semantics encountered by cyclists remains an open question. By benchmarking these specialized models alongside general-purpose models, we not only provide a rigorous assessment of their capabilities, but also establish a foundation for future cyclist-assistive intelligent systems.

\section{Benchmark Construction}
\label{benchmarkconstruction}

We first outline the design principles underlying CyclingVQA, followed by a detailed description of the benchmark tasks (\S~\ref{sec:tasks}) and the question--answer generation pipeline(\S~\ref{sec:qageneration}). \\

\rparagraph{Background.} Our objective is to evaluate the cyclist-assistive traffic scene understanding capabilities of VLMs from a cyclist’s egocentric perspective. To ground the benchmark in real-world traffic regulations, we draw on cycling-specific rulebooks from authoritative sources, including the Deutscher Verkehrssicherheitsrat safety booklet~\cite{dvr2022cycling}, the Vienna Convention on Road Signs and Signals~\cite{vienna_convention_road_signs}, and European urban cycling standards~\cite{hiron2014signs,cycling_in_germany_2016,adfc_traffic_rules_safe_bicycle_rides_2022}.  
These standards specify cyclist-specific guidance for lane usage, right-of-way, and interactions with pedestrians. Notably, cyclist-specific signage (see Table~\ref{tab:cycle_traffic_signs} in the Appendix) follows design principles analogous to standard motor vehicle signage. This is evident, for example, in the consistent use of blue backgrounds for regulatory signs (\includegraphics[height=1em]{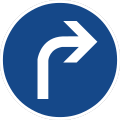} $\rightarrow$ \includegraphics[height=1em]{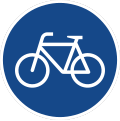}) and red borders for prohibitory signs (\includegraphics[height=1em]{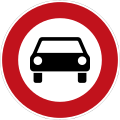} $\rightarrow$ \includegraphics[height=1em]{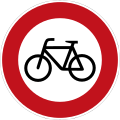}).   This structural consistency, together with the substantial overlap in task demands between cycling and driving, motivates an investigation into whether VLMs specialized for autonomous driving can be applied to cyclist-centric traffic understanding through straightforward prompting. We examine this question through the dedicated benchmark described below.

\begin{figure}[!t]
    \centering
    \begin{minipage}{0.48\linewidth}
        \centering
        \resizebox{\linewidth}{!}{%
            \begin{tabular}{@{}lccccc@{}} 
            \toprule
            \textbf{Task Name} & \textbf{\#Im.} & \textbf{\#Qu.} & \textbf{\#Ch.} & \textbf{Anno.} & \textbf{Markers} \\
            \midrule
            \multicolumn{6}{l}{\textbf{Domain-Independent} \textcolor{cyan}{\faGlobeAmericas}} \\
            \faTrafficLight \ Traffic Sign Grounding & 410 & 410 & 6 & \humanlogo & \textcolor{gray}{\faVectorSquare} Box \\
            \faArrowsAltH \ Spatial Understanding & 153 & 179 & 2 & \humanlogo & \textcolor{gray}{\faVectorSquare} Box \\
            \faHistory \ Temporal Ordering & 110 & 111 & 2 & \humanlogo & -- \\
            \midrule
            \multicolumn{6}{l}{\textbf{Domain-Specific} \textcolor{orange}{\faBicycle}} \\
            \faRoad \ Lane Recognition & 157 & 157 & 2 & \humanlogo & \textcolor{blue}{\faGripLinesVertical} Lane \\
            \faBullseye \ Referring Expr. Det. & 188 & 188 & 2 & \humanlogo & \textcolor{gray}{\faVectorSquare} Box \\
            \faSearchPlus \ Traffic Sign Recog. (w/ B.B) & 259 & 259 & 7 & \chatgptlogo + \humanlogo & \textcolor{gray}{\faVectorSquare} Box \\
            \faIdCard \ Traffic Sign Recog. & 465 & 465 & 7 & \chatgptlogo + \humanlogo & -- \\
            \faRoute \ Sign--Action Assoc. & 225 & 240 & 2 & \chatgptlogo + \humanlogo & \textcolor{blue}{\faGripLinesVertical} Lane \\
            \midrule
            \textbf{Total (All Tasks)} & \textbf{695} & \textbf{2009} & -- & -- & -- \\
            \bottomrule
            \end{tabular}
        }
        \captionof{table}{\textbf{Summary of \textnormal{CyclingVQA} tasks.}}
        \label{tab:datasetstats}
    \end{minipage}
    \hfill
    \begin{minipage}{0.48\linewidth}
        \centering
        \includegraphics[width=\linewidth]{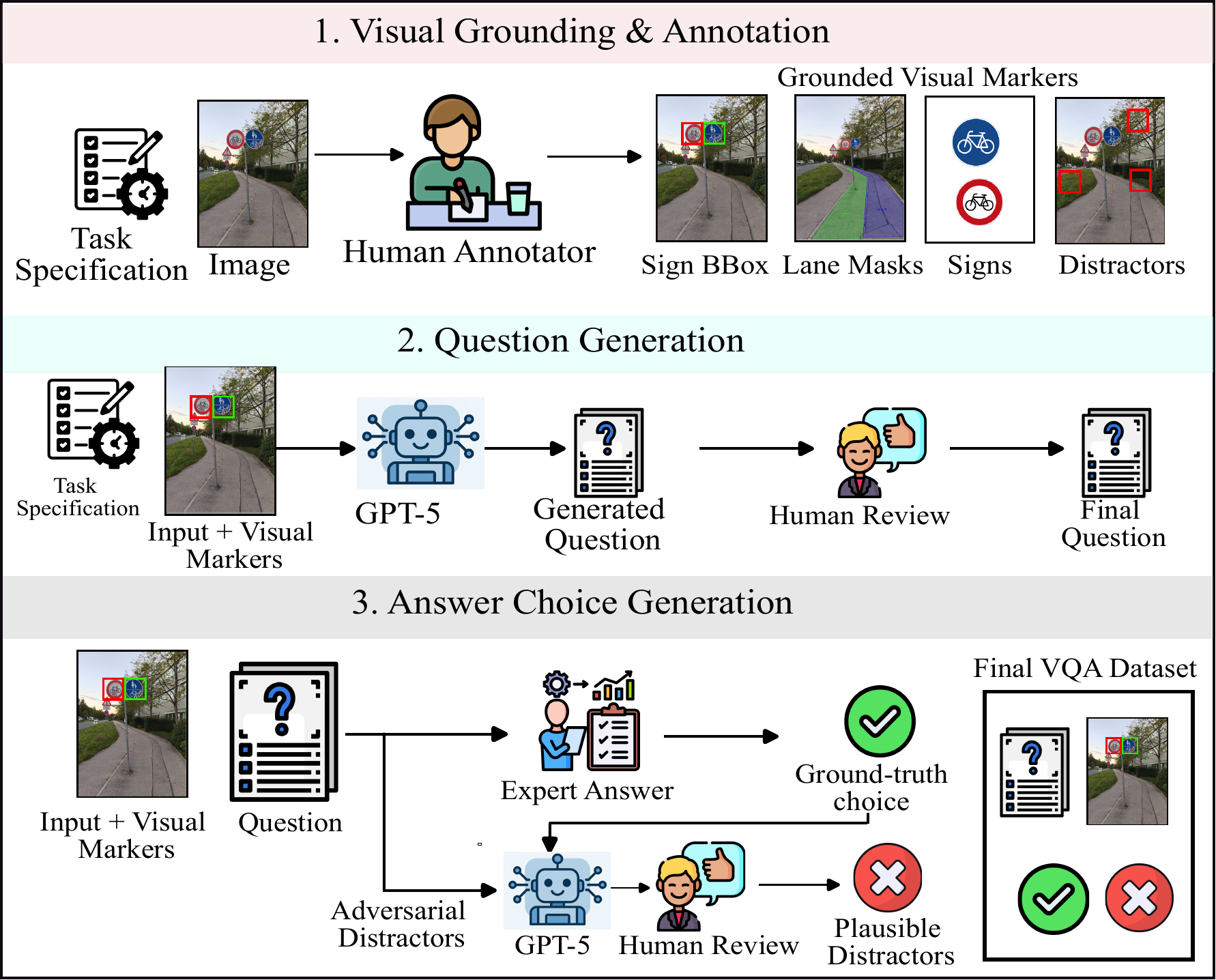}
        \caption{\textbf{Overview of our annotation pipeline.}}
        \label{fig:annotationpipeline}
    \end{minipage}
\end{figure}

\subsection{Benchmark Tasks}\label{sec:tasks}

Inspired by recent fine-grained visual question answering benchmarks~\cite{fu2024blink, danish2025geobench, li2025fine}, we design CyclingVQA, in a similar vein, to evaluate cyclist-centric perception and navigation-relevant reasoning. That is, the benchmark probes whether VLMs can (1) recognize and spatially localize cyclist-relevant traffic elements, and (2) reason about traffic sign--to--lane associations required for safe navigation.

Concretely, CyclingVQA comprises eight task categories: three general visual reasoning tasks and five cyclist-specific domain tasks. Collectively, these tasks assess visual grounding, spatial and temporal understanding, and navigation-relevant decision making from a cyclist’s egocentric perspective. Each task is formulated as an image–question pair and may additionally include visual supervision, such as bounding boxes for traffic signs or pixel-level lane segmentation. Representative examples of all task categories are shown in Figure~\ref{fig:tasks_illustration}.

\noindent\textbf{Domain-Independent Tasks.}  
These tasks evaluate general spatial and temporal understanding and do not require cycling-specific knowledge.

\begin{itemize}
    \item \textbf{Traffic Sign Grounding (TSG):} This task measures the ability to locate traffic signs in an image by selecting the correct bounding box.
    \item \textbf{Spatial Understanding (SU):} This task evaluates understanding of the relative positions of two traffic signs, such as which one is closer or whether one is to the left or right of the other.
    \item \textbf{Temporal Ordering (TO):} This task tests spatio-temporal reasoning by determining the correct order of two nearby frames that contain traffic-related elements.
\end{itemize}

\noindent\textbf{Domain-Specific Tasks.}  
These tasks focus on cyclist-specific perception and reasoning that are important for real-world navigation and safety.

\begin{itemize}
    \item \textbf{Lane Recognition (LR):} This task tests lane-rule association i.e., whether the model can identify the cyclist sign and select the lane intended for cyclists among multiple lane options.
    \item \textbf{Referring Expression Detection (RED):} This task requires matching a textual description of a traffic sign to the correct bounding box in the image.
    \item \textbf{Traffic Sign Recognition with Box Supervision (TSR+S):} Given a single bounding box highlighting a traffic sign, this task evaluates whether the model can choose the correct semantic description of the sign.
    \item \textbf{Traffic Sign Recognition (TSR):} This task removes bounding box supervision and evaluates whether the model can recognize and interpret traffic signs directly from the full image.
    \item \textbf{Sign--Action Association (SAA):} This task evaluates navigation-related reasoning by requiring the model to recognize a traffic sign and decide whether a cyclist is allowed to proceed on a given user-labeled lane.
\end{itemize}

\subsection{QA Generation}\label{sec:qageneration}

We collect \textbf{695} diverse urban street images from a cyclist’s egocentric perspective, covering a wide range of cycling-related traffic scenes and signage. All images are captured at a high resolution of approximately $3060 \times 4080$ pixels to ensure that traffic signs remain legible even at long distances, which is critical for early decision-making. An overview of the benchmark construction pipeline is shown in Figure~\ref{fig:annotationpipeline}.  

Each image is first manually annotated with task-relevant traffic sign bounding boxes and labels, lane segmentations, and relative depth cues. For each task category, we then prompt GPT-5~\cite{gpt5} to generate up to five closed-ended question templates (see Table~\ref{tab:extended_task_questions} in the Appendix) with varied phrasings to reduce prompt sensitivity. Each question is subsequently paired with a single human-annotated ground-truth answer, while distractor options are strategically constructed to challenge the model’s reasoning. For non-binary tasks, GPT-5~\cite{gpt5} is conditioned on the correct answer to generate semantically plausible distractors that require fine-grained scene analysis. Depending on the task, up to six answer choices are included to increase difficulty, and the position of the correct answer is randomized to mitigate positional bias.

Finally, and most importantly, questions are embedded within a system prompt that instructs models to provide both the selected option and \emph{a concise explanation} (see Table~\ref{tab:instruct_prompt} for the prompt). These explanations are used for downstream qualitative analysis of failure modes (\S~\ref{sec:erroranalysis}). In total, \textnormal{CyclingVQA} comprises \textbf{2,053} visual question--answer pairs spanning eight task categories, as summarized in Table~\ref{tab:datasetstats}. Additional statistical studies are provided in Appendix~\ref{sec:datasetanalysis}.

\begin{table*}[t!]
\centering
\renewcommand{\arraystretch}{1.3}
\resizebox{0.98\textwidth}{!}{%
\begin{tabular}{rrrccccccccccc}
\toprule
& & & &
\multicolumn{3}{c}{\textbf{General Tasks}} &
\multicolumn{5}{c}{\textbf{Domain-Specific Tasks}} &
& \\
\cmidrule(lr){5-7}
\cmidrule(lr){8-12}

\textbf{Model} & \textbf{Size} & \textbf{Type} & \textbf{Release} &
\textbf{SU} & \textbf{TSG} & \textbf{TO} & \textbf{TSR+S} & \textbf{RED} & \textbf{TSR} & \textbf{LR} & \textbf{SAA} &
\textbf{Avg} & \textbf{Rank} \\
\midrule
Random & - & - & - & $49.2$ & $16.3$ & $51.4$ & $12.9$ & $42.6$ & $13.5$ & $36.9$ & $51.2$ & $34.3$ & $31$ \\
\midrule
& \multicolumn{10}{c}{\bf Proprietary VLMs} & & \\
\midrule
\rowcolor{teal!15} Gemini-2.5-Flash\cite{comanici2025gemini} & N/A & Reason & 07/2025 & $77.7$ & $\mathbf{98.0}$ & $55.0$ & $82.4$ & $94.7$ & $83.8$ & $\mathbf{72.6}$ & $\mathbf{90.0}$ & $\mathbf{81.8}$ & $1$ \\
\rowcolor{teal!15} GPT-5.1\cite{gpt5} & N/A & Reason & 11/2025 & $63.7$ & $90.0$ & $\mathbf{58.6}$ & $\mathbf{83.0}$ & $94.1$ & $\mathbf{85.3}$ & $59.9$ & $86.2$ & $77.6$ & $2$ \\
\midrule
& \multicolumn{10}{c}{\bf \faGlobeAmericas \ Generalist VLMs} & & \\
\midrule
\rowcolor{orange!15} Eagle2.5-8B\cite{chen2025eagle} & 8B & Instruct & 04/2025 & $53.6$ & $87.8$ & $51.4$ & $50.5$ & $82.4$ & $39.8$ & $53.5$ & $82.5$ & $62.7$ & $12$ \\
\rowcolor{orange!15} InternVL3\cite{zhu2025internvl3} & 8B & Instruct & 04/2025 & $51.4$ & $88.0$ & $49.5$ & $49.5$ & $82.4$ & $48.6$ & $42.0$ & $72.1$ & $60.5$ & $15$ \\
\rowcolor{orange!15} InternVL3.5-2B\cite{wang2025internvl3} & 2B & Instruct & 08/2025 & $62.6$ & $77.6$ & $53.2$ & $47.5$ & $74.5$ & $49.0$ & $42.7$ & $65.8$ & $59.1$ & $17$ \\
\rowcolor{orange!15} InternVL3.5-8B\cite{wang2025internvl3} & 8B & Instruct & 08/2025 & $57.5$ & $88.8$ & $53.2$ & $63.4$ & $87.2$ & $62.9$ & $61.8$ & $78.8$ & $69.2$ & $6$ \\
\rowcolor{orange!15} Molmo2-8B\cite{molmo2} & 8B & Instruct & 12/2025 & $56.4$ & $88.0$ & $51.4$ & $34.2$ & $73.9$ & $37.5$ & $48.4$ & $44.2$ & $54.2$ & $23$ \\
\rowcolor{orange!15} Ovis2.5-2B\cite{lu2025ovis2} & 2B & Instruct & 08/2025 & $72.1$ & $96.3$ & $49.5$ & $63.0$ & $85.1$ & $58.7$ & $45.9$ & $78.3$ & $68.6$ & $7$ \\
\rowcolor{orange!15} Ovis2.5-9B\cite{lu2025ovis2} & 9B & Instruct & 08/2025 & $70.9$ & $97.3$ & $51.4$ & $81.5$ & $93.6$ & $72.2$ & $62.4$ & $79.6$ & $76.1$ & $4$ \\
\rowcolor{orange!15} Phi-4\cite{abouelenin2025phi} & 8B & Instruct & 02/2025 & $44.1$ & $73.7$ & $49.5$ & $59.1$ & $76.6$ & $59.5$ & $50.3$ & $73.3$ & $60.8$ & $14$ \\
\rowcolor{orange!15} Qwen2.5-VL\cite{bai2025qwen2} & 7B & Instruct & 02/2024 & $52.5$ & $81.5$ & $51.4$ & $43.9$ & $71.3$ & $43.2$ & $48.4$ & $77.1$ & $58.7$ & $18$ \\
\rowcolor{orange!15} Qwen3-VL\cite{bai2025qwen3} & 2B & Instruct & 11/2025 & $50.8$ & $97.6$ & $52.3$ & $74.0$ & $84.6$ & $74.1$ & $36.9$ & $72.9$ & $67.9$ & $8$ \\
\rowcolor{orange!15} Qwen3-VL\cite{bai2025qwen3} & 8B & Instruct & 11/2025 & $75.4$ & $89.3$ & $51.4$ & $78.5$ & $\mathbf{95.2}$ & $81.1$ & $58.0$ & $80.8$ & $76.2$ & $3$ \\
\rowcolor{orange!15} LLaVA-Next\cite{liu2024llavanext} & 8B & Instruct & 04/2024 & $44.1$ & $36.1$ & $51.4$ & $25.2$ & $54.3$ & $37.5$ & $27.4$ & $34.2$ & $38.8$ & $28$ \\
\rowcolor{orange!15} LLaVA-OneVision\cite{li2024llava} & 7B & Instruct & 06/2024 & $54.2$ & $65.4$ & $48.6$ & $37.6$ & $66.0$ & $34.0$ & $36.3$ & $71.2$ & $51.7$ & $24$ \\
\rowcolor{orange!15} LLaVA-1.6\cite{liu2024llavanext} & 7B & Instruct & 12/2023 & $46.9$ & $23.2$ & $51.4$ & $26.2$ & $47.9$ & $32.4$ & $15.9$ & $59.6$ & $37.9$ & $29$ \\
\midrule
& \multicolumn{10}{c}{\bf \faVectorSquare \ Spatial-Aware VLMs} & & \\
\midrule
\rowcolor{green!15} PerceptionLM\cite{cho2025perceptionlm} & 3B & Instruct & 04/2025 & $55.9$ & $87.6$ & $51.4$ & $49.7$ & $86.2$ & $66.0$ & $42.7$ & $35.4$ & $59.3$ & $16$ \\
\rowcolor{green!15} PerceptionLM\cite{cho2025perceptionlm} & 8B & Instruct & 04/2025 & $\mathbf{78.8}$ & $95.1$ & $48.6$ & $68.4$ & $86.2$ & $78.8$ & $66.2$ & $58.8$ & $72.6$ & $5$ \\
\rowcolor{green!15} FoundationMotion\cite{bai2025qwen2} & 7B & Instruct & 12/2025 & $49.2$ & $82.4$ & $51.4$ & $43.2$ & $69.7$ & $39.4$ & $59.9$ & $67.5$ & $57.8$ & $19$ \\
\rowcolor{green!15} SenseNova\cite{cai2025scaling} & 8B & Instruct & 10/2025 & $\mathbf{78.8}$ & $70.5$ & $51.4$ & $48.4$ & $83.0$ & $49.4$ & $49.7$ & $68.3$ & $62.4$ & $13$ \\
\rowcolor{green!15} SpatialReasoner\cite{ma2025spatialreasoner} & 7B & Reason & 04/2025 & $37.4$ & $55.1$ & $44.1$ & $33.5$ & $54.3$ & $30.9$ & $57.3$ & $55.8$ & $46.1$ & $27$ \\
\rowcolor{green!15} SpatialThinker\cite{batra2025spatialthinker} & 7B & Reason & 11/2025 & $58.1$ & $95.1$ & $50.5$ & $57.4$ & $87.2$ & $47.9$ & $43.3$ & $71.7$ & $63.9$ & $11$ \\
\rowcolor{green!15} VST\cite{yang2025visual} & 7B & Reason & 11/2025 & $78.2$ & $72.4$ & $51.4$ & $32.5$ & $58.0$ & $30.5$ & $51.0$ & $72.9$ & $55.9$ & $21$ \\
\midrule
& \multicolumn{10}{c}{\bf \faCarSide \ Driving-Centric VLMs } & & \\
\midrule
\rowcolor{blue!15} Cosmos-Reason1\cite{azzolini2025cosmos} & 7B & Reason & 03/2025 & $45.3$ & $61.0$ & $54.1$ & $35.3$ & $64.9$ & $42.5$ & $52.2$ & $79.2$ & $54.3$ & $22$ \\
\rowcolor{blue!15} Cosmos-Reason2\cite{cosmosreason2} & 8B & Reason & 12/2025 & $52.5$ & $79.3$ & $53.2$ & $73.5$ & $87.8$ & $62.9$ & $56.1$ & $70.4$ & $67.0$ & $9$ \\
\rowcolor{blue!15} DriveLMMo1\cite{ishaq2025drivelmm} & 8B & Reason & 03/2025 & $57.5$ & $76.3$ & $50.5$ & $43.2$ & $72.9$ & $46.3$ & $42.7$ & $70.4$ & $57.5$ & $20$ \\
\rowcolor{blue!15} DriveMM\cite{drivemm} & 7B & Instruct & 12/2024 & $54.7$ & $54.1$ & $51.4$ & $30.1$ & $60.6$ & $29.3$ & $45.2$ & $67.5$ & $49.1$ & $26$ \\
\rowcolor{blue!15} ReCogDrive\cite{li2025recogdrive} & 8B & Instruct & 06/2025 & $48.0$ & $50.2$ & $51.4$ & $37.0$ & $54.3$ & $38.6$ & $53.5$ & $64.6$ & $49.7$ & $25$ \\
\rowcolor{blue!15} Dolphins\cite{ma2023dolphins} & 7B & Instruct & 12/2023 & $46.4$ & $15.4$ & $36.9$ & $14.0$ & $44.1$ & $16.2$ & $49.7$ & $70.8$ & $36.7$ & $30$ \\
\bottomrule
\end{tabular}%
}
\caption{\textbf{Evaluation of VLMs on the CyclingVQA benchmark.} We report accuracy (\%) across eight tasks, observing that generalist models outperform driving-centric VLMs overall.}
\label{tab:cyclingvqa_results}
\end{table*}

\section{Experimental Results}\label{sec:expresults}

In Section~\ref{sec:results}, we present the empirical results and summarize the key findings. Section~\ref{sec:cotprompting} then examines the impact of Chain-of-Thought (CoT) prompting. In Section~\ref{sec:erroranalysis}, we analyze the main failure modes, followed by a quantitative analysis of the generated tokens in Section~\ref{sec:quantitativeresults}.\\
 
\subsection{Experimental Setup}

\noindent\textbf{General-Purpose Models.}
We evaluate 16 open-source, general-purpose VLMs (1B--9B parameters) from the Qwen-VL~\cite{bai2025qwen2}, Eagle~\cite{chen2025eagle}, LLaVA-Nxt~\cite{liu2024llavanext}, LLaVA-OneVision~\cite{li2024llava}, InternVL3~\cite{zhu2025internvl3}, Phi~\cite{abouelenin2025phi}, Ovis~\cite{lu2025ovis2}, and Molmo~\cite{molmo2} families, representing widely used baselines in 2025 for multimodal understanding.

\noindent\textbf{Specialist Models.}
To assess domain-specific performance, we include 6 driving-focused VLMs: Cosmos-Reason\{1,2\}~\cite{azzolini2025cosmos,cosmosreason2}, DriveLMM-O1~\cite{ishaq2025drivelmm}, RecogDrive~\cite{li2025recogdrive}, Dolphins~\cite{ma2023dolphins}, and DriveMM~\cite{drivemm}. Additionally, we evaluate 7 spatial-focused specialists, including PerceptionLM~\cite{cho2025perceptionlm}, VST~\cite{yang2025visual}, SpatialThinker~\cite{batra2025spatialthinker}, SenseNova~\cite{cai2025scaling}, and SpatialReasoner~\cite{ma2025spatialreasoner}. 

\noindent \textbf{Proprietary Models.} We also evaluate two frontier models: GPT-5.1~\cite{gpt5} and Gemini-2.5-Flash~\cite{comanici2025gemini}.  

\noindent \textbf{Evaluation Metric.} In all cases, we report accuracy, defined as the percentage of correctly answered questions within each task category. We present both per-task accuracy and the overall mean accuracy averaged across all task categories.

\noindent {\bf Implementation.} We employ Gemma2-9B~\cite{team2024gemma} as a parser (see Table~\ref{tab:parsing_prompt}) to extract discrete answer labels from generated responses, as some VLMs provide textual explanations without explicitly indicating a choice index. For inference, we allocate a maximum generation budget of $4{,}096$ tokens per query. Additional implementation details are provided in Appendix~\ref{sec:implementation}.

\subsection{Main Results}\label{sec:results}
Table~\ref{tab:cyclingvqa_results} summarizes performance across the eight \textnormal{CyclingVQA} task categories. The proprietary model Gemini-2.5-Flash achieves the highest overall accuracy; several consistent trends from our evaluation are distilled below.

\paragraph{Generalist VLMs outperform driving-specialized models.}
Surprisingly, despite being fine-tuned on traffic environments, driving-specialized models consistently lag behind strong generalist VLMs on CyclingVQA. For example, specialist instruct models such as \textnormal{Dolphins} (36.7\%) and \textnormal{DriveMM} (49.1\%) perform substantially worse than generalist baselines on domain-specific tasks. Even reasoning-based driving models, including the latest \textnormal{Cosmos-Reason2} (67.0\%) and \textnormal{DriveLMM-O1} (57.5\%), are surpassed by generalist architectures such as \textnormal{Qwen3-VL-8B} (76.2\%) and \textnormal{Ovis2.5-9B} (76.1\%). These results, while unexpected, points to  weak generalization capabilities of driving VLMs. 

\paragraph{Instruct models outperform reasoning models.}
Across the six reasoning models evaluated, performance consistently trails that of instruction-following models. The strongest reasoning model, \textnormal{Cosmos-Reason2} (67.0\%), ranks only ninth overall. In stark contrast, token-efficient instruct models dominate the leaderboard, with \textnormal{Qwen3-VL-8B} (76.2\%) ranking third and achieving the best overall performance among open-source VLMs.

\paragraph{Model scale is not a reliable predictor of performance.}
We do observe that larger models do not necessarily outperform smaller ones on \textnormal{CyclingVQA}. Several recent compact models achieve performance comparable to—or exceeding—that of much larger older counterparts. For instance, \textnormal{Ovis2.5-2B} (68.6\%) and \textnormal{Qwen3-VL-2B} (67.9\%) outperform larger models such as \textnormal{Qwen2.5-7B} (58.7\%) and \textnormal{InternVL3-8B} (60.5\%). We hypothesize that superior training data quality in more recent models may have played an important role in this behavior.

\paragraph{Driving-specialized models struggle with spatial understanding.}
Another interesting finding is that driving-centric models show particular weaknesses on spatial reasoning tasks. On the SU task, the best-performing driving-specialized model, \textnormal{DriveLMM-O1}, achieves only 57.5\% accuracy, barely exceeding the random baseline of 50\%. In contrast, generalist models such as \textnormal{Qwen3-VL-8B} reach 75.4\%, while spatially enhanced architectures perform even better, with \textnormal{PerceptionLM-8B} achieving the highest accuracy of 78.8\%.

\paragraph{Temporal reasoning remains challenging.}
Temporal reasoning emerges as the most difficult task in our benchmark. Even frontier models such as \textnormal{Gemini-2.5-Flash} achieve only 58.6\% accuracy, again only marginally above the random baseline. This result underscores that current models lack the ability to consistently pivot their reasoning around visible traffic signs across frames. Indeed, these findings warrant deeper investigation into the limitations of current VLMs in physical and temporal understanding.

\begin{table*}[t]
    \centering
    \vspace{2mm} %
    \includegraphics[width=0.99\linewidth]{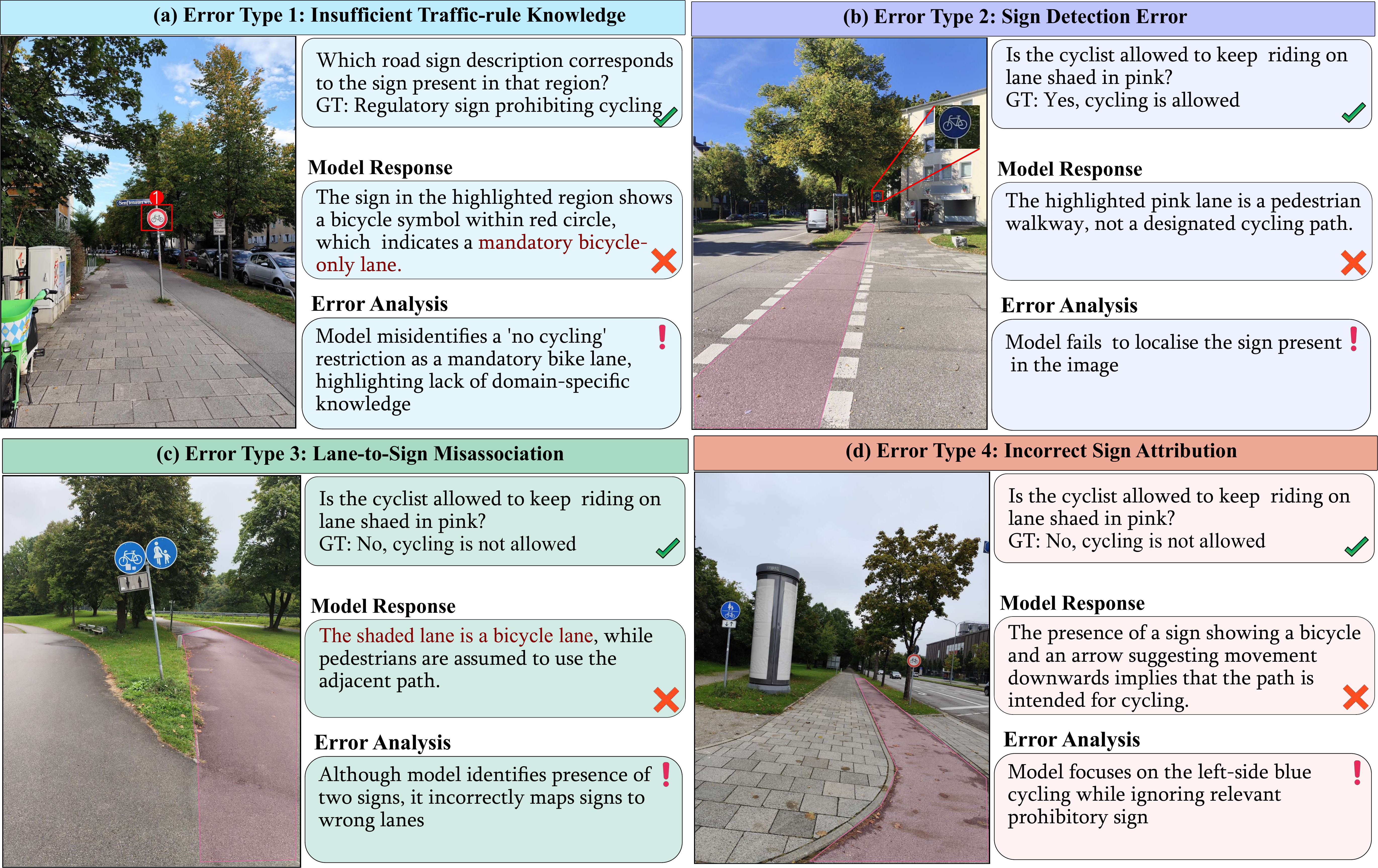}
    \caption{{\bf Taxonomy of Failure Modes.} We characterize model errors across four recurring categories, providing a systematic overview of current VLM limitations in cyclist-centric scenarios. See Appendix~\ref{sec:qualitativeresults} for a comprehensive qualitative analysis and additional case studies.}
    \label{tab:failure_error_categories}
\end{table*}

\begin{figure}[!t]
    \centering
    \includegraphics[width=0.95\linewidth]{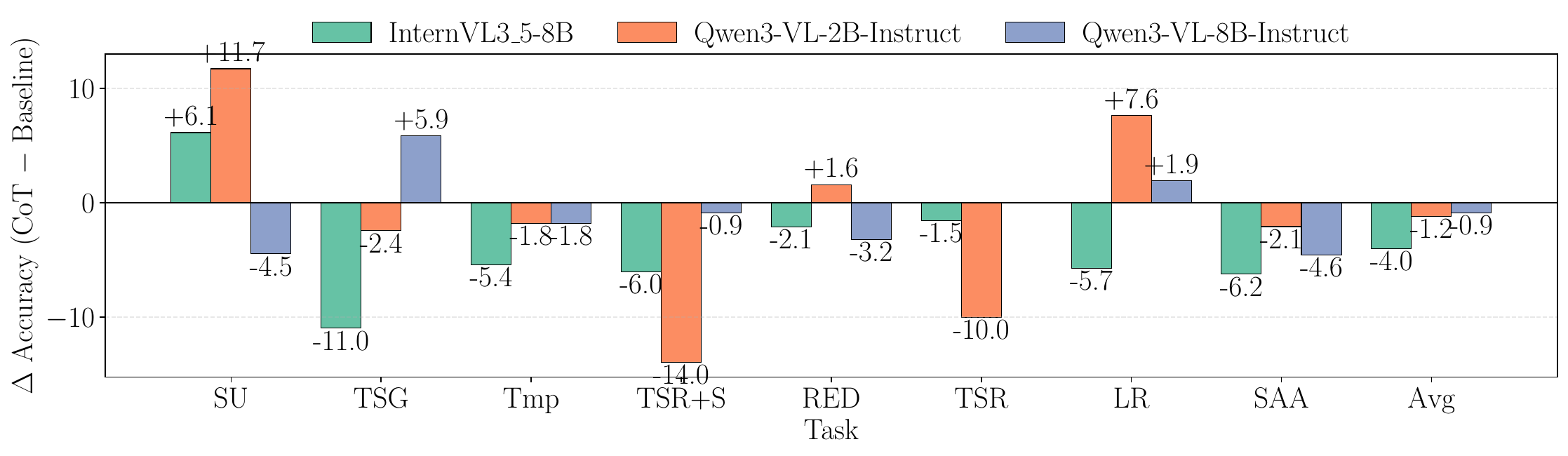}
\caption{{\bf CoT vs. Standard Prompting.} Overall performance degrades under CoT prompting across the three instruct models.}
\label{fig:cotimpact}
\end{figure}

\subsection{Impact of CoT-Prompting}\label{sec:cotprompting}
We examine the effect of explicit CoT prompting (See Table~\ref{tab:cot_prompt}) by comparing baseline and CoT performance on three instruct models. As shown in Figure~\ref{fig:cotimpact}, CoT prompting does not lead to consistent improvements and instead results in average accuracy drops of 4.0\%, 1.2\%, and 0.9\% for \textnormal{InternVL3.5-8B}, \textnormal{Qwen3-VL-2B}, and \textnormal{Qwen3-VL-8B}, respectively. Although CoT yields occasional gains on certain spatial reasoning tasks, these improvements are outweighed by performance degradation on grounding and association tasks.  Overall, the predominantly negative deltas suggest that, for cycling-specific VQA, direct instruction-following is more reliable than explicit multi-step reasoning with the current models; however, we emphasize that this finding is not conclusive and may reflect limitations of current prompting strategies rather than an inherent weakness of CoT reasoning.

\subsection{Qualitative Analysis of Failure Modes}\label{sec:erroranalysis}

While quantitative results are informative, they do not tell the whole story. We therefore turn to a more pressing question: what are the key failure modes that limit cyclist-assistive reasoning in current VLMs? To this end, we identify four recurring error modes related to spatial reasoning and safety-critical decision-making. Table~\ref{tab:failure_error_categories} summarizes these failure modes, with additional qualitative examples in Appendix~\ref{sec:qualitativeresults}.

\noindent\textbf{1. Misinterpretation of traffic rule semantics.}
A frequent source of error occurs when models incorrectly interpret the semantics of traffic signs. While signs are often detected and localized correctly, their regulatory meaning is misclassified, resulting in unsafe or invalid action predictions. For instance, prohibitory signs are sometimes mistaken for permissive or mandatory indicators (Table~\ref{tab:failure_error_categories}(a)). These failures suggest that translating visual sign cues into cyclist-specific traffic rules remains unreliable. 

\noindent\textbf{2. Perception failures.}
Models also exhibit failures at the perception stage, particularly in visually cluttered environments or when signs are small, distant, or partially occluded (Table~\ref{tab:failure_error_categories}(b)). When critical signage is missed, models often fall back on contextual priors, occasionally hallucinating cues that are not supported by the visual evidence. Such behavior leads to confident but incorrect decisions that ignore safety-relevant constraints.

\noindent\textbf{3. Lane--sign association errors.}
Even when traffic signs are correctly detected, models frequently struggle to associate them with the appropriate road lane. This issue is most pronounced at intersections or shared-use paths where multiple, potentially conflicting, signs appear in close proximity (Table~\ref{tab:failure_error_categories}(c)). In these cases, models may recognize all signs but fail to determine which applies to the cyclist’s trajectory, revealing weaknesses in spatial association. Moreover, models tend to rely on the physical location of traffic poles, leading to incorrect lane associations (see Table~\ref{tab:casestudy3}).

\noindent\textbf{4. Incorrect attribution of relevant signage.}
Finally, in scenes containing multiple visible signs, models sometimes base decisions on irrelevant signage while overlooking the sign that directly governs the cyclist’s path (Table~\ref{tab:failure_error_categories}(d)). This failure mode suggests that models struggle to correctly rank the relevance of multiple signs, frequently conditioning their decisions on cues that do not apply to the cyclist’s intended trajectory.

\subsection{Generation Verbosity vs.\ Performance}\label{sec:quantitativeresults}

{\noindent \bf Generation verbosity is not strongly correlated with performance.}
Figure~\ref{fig:generated_answer_lengths} reports the average number of generated tokens per datapoint for each model and task. To account for differences in tokenization across VLMs, we tokenize all generated outputs using the GPT-2 tokenizer~\cite{radford2019language}. Among models explicitly optimized for long-form reasoning, Cosmos-Reason1~\cite{azzolini2025cosmos} produces the highest average of $546$ tokens per response, yet ranks only 21st on the leaderboard in terms of overall task performance. In contrast, Qwen3-VL-8B achieves the top rank among open-source VLMs while generating just $50$ tokens on average. These results suggest that increased generation verbosity does not necessarily translate to improved performance.

\begin{figure*}[t]
\centering
\includegraphics[width=0.99\textwidth]{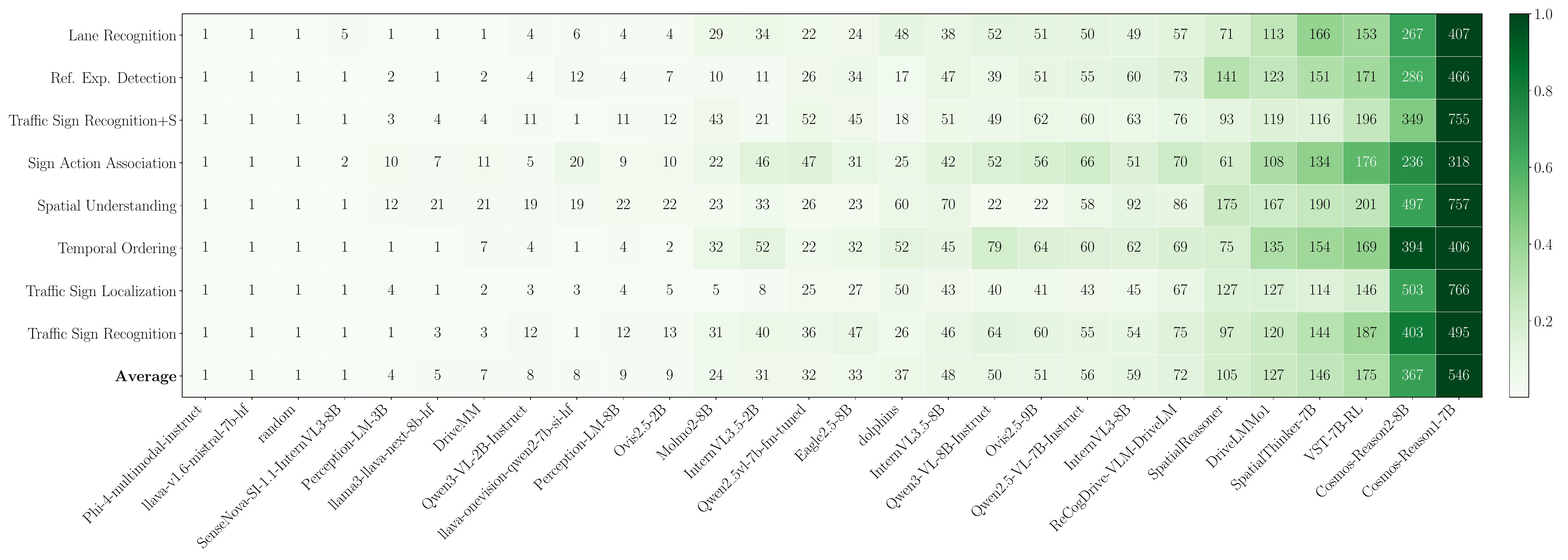}
\caption{{\bf Characterizing Generation Verbosity.} We report the mean number of tokens generated per response. Recall that our standard prompting setup (see Table~\ref{tab:instruct_prompt}) elicits a dual-part response: an initial selection followed by a brief reasoning.}
\label{fig:generated_answer_lengths}\label{fig:generated_answer_lengths}
\end{figure*}

\section{Conclusion}\label{sec:conclusion}

\noindent In this work, we introduce \textnormal{CyclingVQA}, a benchmark constructed from real-world traffic scenes captured from a cyclist’s egocentric perspective to evaluate cyclist-assistive capabilities of VLMs. The benchmark comprises a diverse set of tasks that probe perception, spatial reasoning, and navigation-relevant decision making critical for safe cycling. Our evaluation across a broad range of VLMs reveals promising capabilities, while also highlighting clear room for improvement in cyclist-centric understanding. In particular, we find that VLMs specialized for autonomous driving exhibit limited transferability to cyclist-perspective scenes, despite shared traffic semantics. Finally, we conduct a manual analysis of representative failures to pinpoint recurring limitations. We hope that CyclingVQA serves as a valuable resource for the community and fosters future research on cyclist-assistive intelligent systems.

\rparagraph{ Ethical Considerations.} Our dataset primarily comprises images of bicycle lanes and cycling infrastructure, thereby minimizing intrusion into personal or private spaces. All data were collected by the authors on non-busy days to further reduce the likelihood of capturing identifiable individuals. The data collection process strictly adheres to GDPR requirements and is conducted solely for academic research in AI safety.
To protect individual privacy, we manually identified and localized faces and vehicle license plates, dedicating over 12 person-hours to this process, and applied Gaussian blurring to all identifiable regions.  \\

\noindent{\bf Disclosure of Funding.} The authors declare no competing interests. This research was self-funded by the authors. No external financial support was received for the dataset collection or for the experimental evaluation. \\

\noindent{\bf LLM Usage.} We disclose the use of ChatGPT~\cite{gpt5} for writing assistance. Specifically, ChatGPT was used to correct grammatical errors and improve sentence flow, with all outputs subsequently reviewed and refined by the authors. We emphasize that the benchmark design, evaluation methodology, and error analysis are entirely the work of the human authors. 

\section{Limitations and Future Work}\label{sec:limitations}

While CyclingVQA provides a first step toward cyclist-centric evaluation, it comes with several limitations that also point to promising directions for future research.

\noindent\textbf{Dataset Scale.}
Compared to large-scale autonomous driving benchmarks~\cite{cao2024maplm,ishaq2025drivelmm,wei2025driveqa,ghosh2025roadwork,tian2025nuscenes,li2025fine}, our dataset may be considered relatively modest in scale. This is largely due to the substantial manual effort required to collect egocentric cyclist data, accurately annotate traffic signs, lane segments, and fine-grained spatial cues, and ensure appropriate privacy safeguards. Scaling the dataset through automated labeling, while preserving annotation quality, therefore remains an important goal for future work.

\noindent\textbf{Dataset Diversity.}
Data collection was conducted predominantly under low-traffic conditions, primarily due to privacy considerations. As a result, the dataset contains fewer instances of dense interactions involving other cyclists, pedestrians, or motor vehicles. We nevertheless include few scenarios in which cyclist intentionally navigate into pedestrian lanes to reflect unsafe behavior. Looking ahead, a broader and more varied set of such unsafe scenarios through generative models could further improve the benchmark diversity. 

\noindent\textbf{Geographical Coverage.}
All data are sourced from Munich, a city known for its highly structured cycling infrastructure~\cite{cleancitiescampaign_2025_cycling_infrastructure}. We acknowledge that it may not fully capture the diversity of cycling environments worldwide, such as rural roads, regions with informal traffic norms, or cities with less developed infrastructure. Expanding geographic coverage is therefore imperative for future work.

\noindent\textbf{Evaluation Framework.}
Our evaluation currently focuses on objective accuracy over closed-ended choices and does not include a quantitative assessment of model reasoning traces. Although the manual error analysis in Section~\ref{sec:erroranalysis} offers qualitative insights, future work could explore more fine-grained metrics for evaluating the correctness and usefulness of generated explanations. In addition, while our initial experiments with chain-of-thought prompting (Section~\ref{sec:cotprompting}) did not yield improvements, a more systematic investigation of prompt sensitivity remains an open direction.

\noindent\textbf{Beyond VLMs.}
Finally, our study focuses exclusively on VLMs. Recent Vision–Language–Action (VLA) models have shown promising capabilities in directly predicting navigation actions~\cite{zhou2025opendrivevla,jiang2025survey,hu2025vision} in autonomous driving scenarios. Extending cyclist-assistive evaluation to such models represents an exciting avenue for future research.

\bibliography{custom}
\bibliographystyle{iclr2026_conference}

\appendix

\section{Appendix Summary}
In the following sections, we provide additional details, experiments, and visualizations to supplement our main findings. \\

\noindent \textbf{Prompt Details (Section~\ref{sec:prompts}):} We provide the system prompts used in our benchmarks, including the various prompt templates and judge templates. \\

\noindent \textbf{Implementation Details (Section~\ref{sec:implementation}):} We outline the experimental implementation details, such as exact model cards and model sources.\\

\noindent \textbf{Additional Quantitative Results (Section~\ref{sec:quantitativeresults}):} We present further experimental insights regarding the tokens generated by different models. \\

\noindent \textbf{Additional Qualitative Results (Section~\ref{sec:qualitativeresults}):} We provide qualitative results through additional case studies and detailed output analysis. \\

\section{Prompt Details}\label{sec:prompts}
This section describes the prompts and parsing strategies used across our experiments.\\

\noindent \textbf{Prompting for Instruct Models:}
Table~\ref{tab:instruct_prompt} lists the prompts used for instruct models (e.g., OVis2.5-8B~\cite{lu2025ovis2}). The prompt instructs the models to return a choice followed by a brief justification. We find that this format is computationally efficient and facilitates interpretation of model decisions; however, some instruct models occasionally omit the justification despite these explicit instructions.\\

\noindent \textbf{Prompting for Reasoning Models:} Table~\ref{tab:think_prompt} shows the prompts used for reasoning-oriented models (e.g., Cosmos-Reason1~\cite{azzolini2025cosmos}). For these models, we do not explicitly request intermediate reasoning, as it is intrinsic to their generation behavior.\\

\noindent \textbf{CoT Prompting Ablation:} As an ablation, we also evaluate Chain-of-Thought (CoT) prompting for instruct models (Table~\ref{tab:cot_prompt}), which requires a step-by-step analysis before providing the final answer. \\

\noindent \textbf{Response Parsing Strategy:} To reduce parsing errors arising due to inconsistent formatting outputs, we use a dedicated choice-parsing model. Given the question, candidate options, and the raw model-generated output, the parsing model extracts a standardized final choice and the corresponding rationale. The parsing prompt is provided in Table~\ref{tab:parsing_prompt}. \\

\noindent
\begin{tcolorbox}[
  colback=white, colframe=black, title=Standard Prompt for Instruct Models, 
  width=\linewidth, enhanced, breakable
]

\begin{lstlisting}[basicstyle=\ttfamily\footnotesize, frame=none, breaklines=true]
Answer the following multiple-choice question by selecting exactly one option.

Question: {question}
Options: {options}

Provide only the selected option letter, followed by a brief reasoning.
\end{lstlisting}
\end{tcolorbox}
\captionof{table}{The instruction template used for instruct-tuned models.}
\label{tab:instruct_prompt}

\vspace{2em}

\noindent
\begin{tcolorbox}[
  colback=white, colframe=black, title=Standard Prompt for Reasoning Models, 
  width=\linewidth, enhanced, breakable
]
\begin{lstlisting}[basicstyle=\ttfamily\footnotesize, frame=none, breaklines=true]
Answer the following multiple-choice question by selecting exactly one option.

Question: {question}
Options: {options}
\end{lstlisting}
\end{tcolorbox}
\captionof{table}{The instruction template used for reasoning-focused models.}
\label{tab:think_prompt}

\vspace{2em}

\noindent
\begin{tcolorbox}[
  colback=white, colframe=black, title=CoT Prompt for Instruct Models, 
  width=\linewidth, enhanced, breakable
]
\begin{lstlisting}[basicstyle=\ttfamily\footnotesize, frame=none, breaklines=true]
Answer the following multiple-choice question by selecting exactly one option.

Question: {question}
Options: {options}

Let's think step by step first and then return the choice.
\end{lstlisting}
\end{tcolorbox}
\captionof{table}{The instruction template used for the Chain-of-Thought (CoT) ablation.}
\label{tab:cot_prompt}

\vspace{2em}

\noindent
\begin{tcolorbox}[
  colback=white, colframe=black, title=Parsing Prompt for Choice Extraction, 
  width=\linewidth, enhanced, breakable
]
\begin{lstlisting}[basicstyle=\ttfamily\footnotesize, frame=none, breaklines=true]
You are a choice-parsing language model.
Your task is to infer the final predicted choice from the generated text.

Rules:
- Valid choices: A, B, C, D, E, F, G, H
- If a clear final choice is stated, extract it.
- If multiple choices appear, select the FINAL one.
- If no choice is explicitly stated, infer the most confident option.
- Return ONLY a valid JSON object.

Question: {question}
Options: {options}
Generated Text: {generated_text}

{
  "predicted_choice": <A/B/.../null>,
  "predicted_reasoning": "Explanation"
}
\end{lstlisting}
\end{tcolorbox}
\captionof{table}{The instruction template used for the choice-parsing model.}
\label{tab:parsing_prompt}

\section{Implementation}\label{sec:implementation}

\noindent {\bf Target VLMs.} We use the opensource models present on the HuggingFace~\cite{wolf-etal-2020-transformers} or respective official GitHub sources for our experiments. We report details about the model card in Table~\ref{tab:model_inventory_detailed}. For reasoning models (eg., Cosmos-Reason2~\cite{azzolini2025cosmos}), we use a budget of $4096$ tokens during generation and use greedy-decoding. 

\begin{table}[h]
\centering
\resizebox{0.6\columnwidth}{!}{%
\begin{tabular}{lll}
\toprule
\textbf{Model} & \textbf{Scale} & \textbf{Hugging Face Model ID} \\
\midrule
Qwen3-VL & 2B & {\ttfamily Qwen/Qwen3-VL-2B-Instruct} \\
Qwen3-VL & 8B & {\ttfamily Qwen/Qwen3-VL-8B-Instruct} \\
Ovis2.5 & 2B & {\ttfamily AIDC-AI/Ovis2.5-2B} \\
Ovis2.5 & 9B & {\ttfamily AIDC-AI/Ovis2.5-9B} \\
InternVL3.5 & 2B & {\ttfamily OpenGVLab/InternVL3.5-2B} \\
InternVL3.5 & 8B & {\ttfamily OpenGVLab/InternVL3.5-8B} \\
Eagle2.5 & 8B & {\ttfamily NExT-GPT/Eagle2.5-8B} \\
Phi-4 & 6B & {\ttfamily microsoft/phi-4} \\
InternVL3 & 8B & {\ttfamily OpenGVLab/InternVL3-8B} \\
Molmo2 & 8B & {\ttfamily allenai/Molmo-7B} \\
FoundationMotion & 7B & {\ttfamily WoWolf/Qwen2\_5vl-7b-fm-tuned} \\
LLaVA-OV & 7B & {\ttfamily llava-hf/llava-onevision-7b-ov} \\
LLaVA-Next & 8B & {\ttfamily llava-hf/llava-next-8b} \\
LLaVA-1.6 & 7B & {\ttfamily liuhaotian/llava-v1.6-7b} \\
\midrule
PerceptionLM & 2B & {\ttfamily facebook/Perception-LM-2B} \\
PerceptionLM & 8B & {\ttfamily facebook/Perception-LM-8B} \\
Qwen2.5-VL & 7B & {\ttfamily Qwen/Qwen2.5-VL-7B} \\
SenseNova & 8B & {\ttfamily sensenova/SenseNova-SI-1.1} \\
SpatialThinker & 7B & {\ttfamily OX-PIXL/SpatialThinker-7B} \\
VST & 7B & {\ttfamily rayruiyang/VST-7B-RL} \\
SpatialReasoner & 7B & {\ttfamily ccvl/SpatialReasoner} \\
\midrule
Cosmos-Reason2 & 8B & {\ttfamily nvidia/Cosmos-Reason2-8B} \\
Cosmos-Reason1 & 7B & {\ttfamily nvidia/Cosmos-Reason1-7B} \\
DriveLMM-O1 & 7B & {\ttfamily ayeshaishaq/DriveLMMo1} \\
DriveMM & 7B & {\ttfamily DriveMM/DriveMM} \\
ReCogDrive & 7B & {\ttfamily owl10/ReCogDrive-VLM} \\
Dolphins & 7B & {\ttfamily github.com/SaFo-Lab/Dolphins} \\
\bottomrule
\end{tabular}%
}
\caption{\textbf{Target VLMs} evaluated in our study.}
\label{tab:model_inventory_detailed}
\end{table}

\begin{figure}[h]
    \centering
    \setlength{\abovecaptionskip}{0cm}
    \includegraphics[width=0.48\columnwidth]{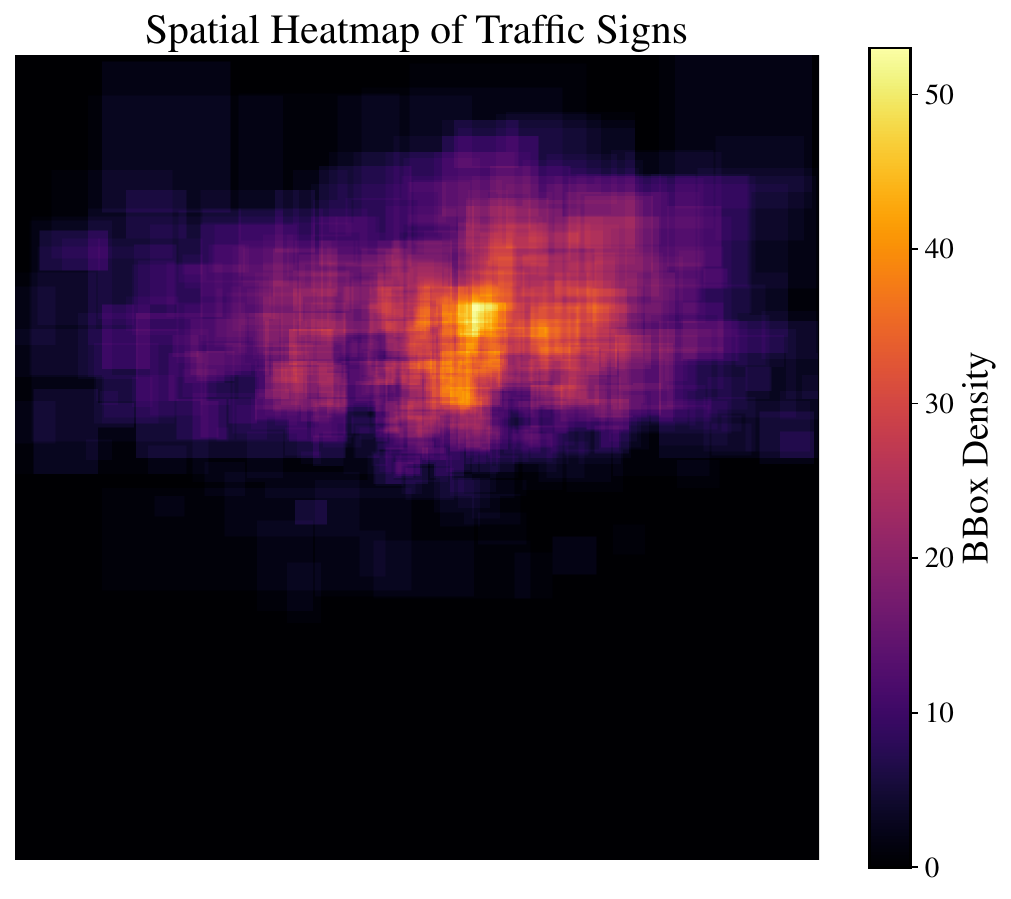}
    \hfill
    \includegraphics[width=0.48\columnwidth]{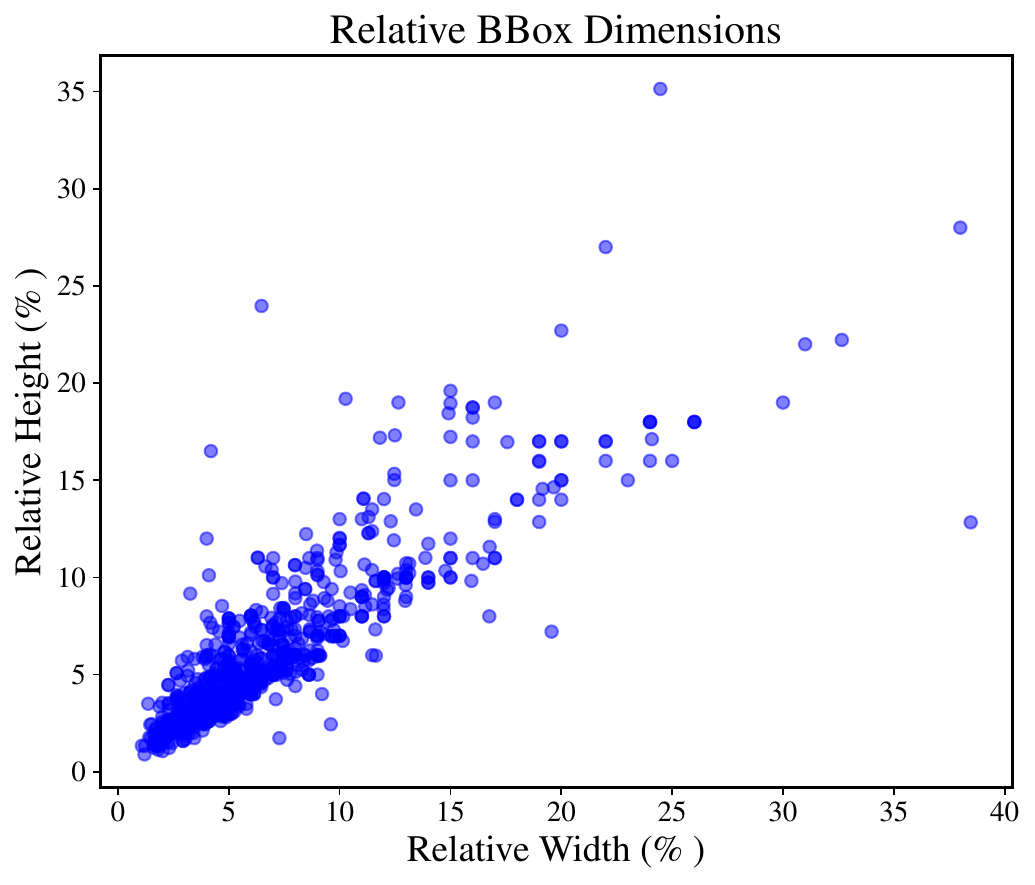}
    \caption{
    {\bf Traffic sign bounding-box statistics.} Left: spatial heatmap of traffic-sign locations. Right: distribution of relative bounding-box widths and heights.
    }
    \label{fig:bbox_sign_stats}
\end{figure}

\section{Additional Details of the Benchmark Dataset}
\label{sec:datasetanalysis}

We visualize the spatial heatmap of traffic sign locations in Figure~\ref{fig:bbox_sign_stats}(a) and the relative area distribution of traffic signs in Figure~\ref{fig:bbox_sign_stats}(b). We  also provide representative sample question templates used in our benchmark in Table~\ref{tab:extended_task_questions}.

\section{Additional Qualitative Results}\label{sec:qualitativeresults}

We present several case studies showing recurring errors for different tasks. These are shown in Tables~\ref{tab:casestudy1}, \ref{tab:casestudy2}, \ref{tab:casestudy3}, \ref{tab:casestudy4}, \ref{tab:casestudy5}, \ref{tab:casestudy6}, \ref{tab:casestudy7}, \ref{tab:casestudy8}, \ref{tab:casestudy9}, \ref{tab:casestudy10}, \ref{tab:casestudy11}, \ref{tab:casestudy12}, \ref{tab:casestudy13}, \ref{tab:casestudy14}, \ref{tab:casestudy15}, \ref{tab:casestudy16}, and \ref{tab:casestudy17}. Note that, the generated text shown in these Figures are automatially paraphrased for brevity using GPT-5~\cite{gpt5}. We refer the reader to the project website for the full evaluation logs.

\begin{table*}[htbp]
\centering

\scriptsize 
\setlength{\tabcolsep}{4pt} 
\begin{tabularx}{\linewidth}{@{} X @{}}
\toprule
\textbf{Questions} \\
\midrule

\rowcolor{gray!10} \textbf{Traffic Sign Localization} \\
\begin{itemize}[nosep,leftmargin=12pt]
  \item    "In this image, the regions <1>, <2>, <3>, <4>, <5>, <6>, <7> are highlighted. Which highlighted region corresponds to a traffic sign?",
  \item    "From the cyclist's perspective, several regions are marked in the image as <1>, <2>, <3>, <4>, <5>, <6>, <7>. Which region actually contains a traffic sign?",
  \item    "From the cyclist's perspective, given the marked regions <1>, <2>, <3>, <4>, <5>, <6>, <7>, <8>, which one localizes a real traffic sign?",
 \item     "From the cyclist's viewpoint, among the marked regions <1>, <2>, <3>, <4>, <5>, <6>, <7>, which region correctly points to a traffic sign?",
  \item    "Looking at the highlighted regions <1>, <2>, <3>, <4>, <5>, <6>, which one contains the traffic sign?"
\end{itemize} \\

\rowcolor{gray!10} \textbf{Temporal Ordering} \\
\begin{itemize}[nosep,leftmargin=12pt]
 \item "From the cyclist’s perspective while riding forward, which of the two images represents an earlier viewpoint?",
 \item   "As the cyclist progressed along the road, which of these two views corresponds to the earlier moment?",
 \item   "While the cyclist was moving forward, two snapshots were captured moments apart. Which image shows the earlier point in the journey?",
 \item   "During continuous forward motion, the cyclist took these two photos. Which one was captured first in time?",
 \item   "These two images were captured sequentially during the cyclist's forward travel. Which image was taken first?"
\end{itemize} \\ \addlinespace

\rowcolor{gray!10} \textbf{Spatial Scene Understanding} \\
\begin{itemize}[nosep,leftmargin=12pt]

\item "In the image, captured from the cyclist’s perspective, focus on the regions marked as <region1> (highlighted by a green bounding box) and <region2> (highlighted by a red bounding box). Describe the spatial relationship between the object in <region1> and the object in <region2>.",
   
\item "In the image, captured from the cyclist’s perspective, focus on the regions marked as <region1> (highlighted by a green bounding box) and <region2> (highlighted by a red bounding box). How is the object in <region1> positioned or related to the object in <region2>?",
   
\item "In the image, captured from the cyclist’s perspective, focus on the regions marked as <region1> (highlighted by a green bounding box) and <region2> (highlighted by a red bounding box). What is the relationship between the object in <region1> and the object in <region2>?"

\end{itemize} \\ \addlinespace

\rowcolor{gray!10} \textbf{Bicycle Lane Prediction} \\
\begin{itemize}[nosep,leftmargin=12pt]
   \item  "From the cyclist’s perspective while riding forward, which of the two images represents an earlier viewpoint?",
  \item    "As the cyclist progressed along the road, which of these two views corresponds to the earlier moment?",
  \item    "While the cyclist was moving forward, two snapshots were captured moments apart. Which image shows the earlier point in the journey?",
  \item    "During continuous forward motion, the cyclist took these two photos. Which one was captured first in time?",
  \item    "These two images were captured sequentially during the cyclist's forward travel. Which image was taken first?"
\end{itemize} \\ \addlinespace

\rowcolor{gray!10} \textbf{Sign--Action Association} \\
\begin{itemize}[nosep,leftmargin=12pt]
    \item "The image shows a forward-facing view from a cyclist’s perspective. Based on the shaded path (in pink) shown ahead, should the cyclist proceed forward on this lane? The shading is externally drawn on the image as an overlay.",
   \item   "The image shows a forward-facing view from a cyclist’s perspective. Based on the shaded route (in pink) ahead, should the cyclist enter this street? The shading is externally drawn on the image as an overlay.",
   \item   "The image shows a forward-facing view from a cyclist’s perspective. From the cyclist’s forward view, can one ride into the shaded street area (in pink)? The shading is externally drawn on the image as an overlay.",
   \item   "The image shows a forward-facing view from a cyclist’s perspective. From the cyclist’s forward view, is cycling allowed along the highlighted lane (in pink)? The highlight is externally drawn on the image as an overlay."
\end{itemize} \\ \addlinespace

\rowcolor{gray!10} \textbf{Traffic Sign Recognition (w/ B.B supervision)} \\
\begin{itemize}[nosep,leftmargin=12pt]

    \item "In the image, focus on the region marked as <1> (highlighted by a red bounding box). Which traffic sign description most accurately refers to the sign visible in that region?",
    \item "In the image, focus on the region marked as <1> (highlighted by a red bounding box). Which specific traffic sign description correctly matches the sign visible in that region?",
    \item "In the image, focus on the region marked as <1> (highlighted by a red bounding box). What is the correct description of the traffic sign visible in that region?",
    \item "In the image, focus on the region marked as <1> (highlighted by a red bounding box). Which official traffic sign description best represents the sign that appears in that region?",
    \item  "In the image, focus on the region marked as <1> (highlighted by a red bounding box). Which road sign description corresponds to the sign present in that region?"

    \end{itemize} \\ \addlinespace

\rowcolor{gray!10} \textbf{Traffic Sign Recognition} \\
\begin{itemize}[nosep,leftmargin=12pt]
    \item     "In the image captured from the cyclist’s forward viewpoint, which single traffic sign pole from the given below options explictly appears along the route ahead?"

\end{itemize} \\ \addlinespace

\rowcolor{gray!10} \textbf{Refering Expression Detection} \\
\begin{itemize}[nosep,leftmargin=12pt]
    \item         "From the cyclist's perspective, focus on the bounding box  regions marked <1>, <2>, <3>. Which region is being referred to as \{"Regulatory sign indicating a mandatory bicycle-only lane ?"\},

\end{itemize} \\ \addlinespace

\bottomrule
\end{tabularx}
\caption{{\bf Sample question templates part of our benchmark.}}
\label{tab:extended_task_questions}
\end{table*}

\begin{table*}[t]
\centering
\renewcommand{\arraystretch}{1.2} %
\setlength{\tabcolsep}{6pt}       %
\begin{tabular}{@{}m{0.15\textwidth} m{0.78\textwidth}@{}}  
\toprule
\textbf{Traffic Sign(s)} & \textbf{Description} \\ 
\midrule

\centering\includegraphics[width=0.09\textwidth]{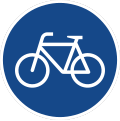} &
\textbf{Sign 237 –} Mandatory cycle path. Indicates a dedicated lane or path reserved exclusively for cyclists. \\
\midrule

\centering\includegraphics[width=0.09\textwidth]{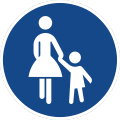} &
\textbf{Sign 239 –} Pedestrian path. Cycling is not permitted on this path. \\
\midrule

\centering\includegraphics[width=0.09\textwidth]{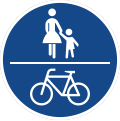} &
\textbf{Sign 240 –} Shared path for pedestrians and cyclists. Both users share the same lane. \\
\midrule

\centering\includegraphics[width=0.09\textwidth]{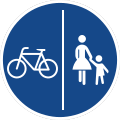} &
\textbf{Sign 241-30 –} Segregated pedestrian and cycle path. The path is divided into separate lanes for each. \\
\midrule

\centering\includegraphics[width=0.09\textwidth]{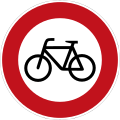} &
\textbf{Sign 254 –} No cycling permitted. Indicates that cyclists are prohibited from entering or using this section. \\
\midrule

\centering\includegraphics[width=0.09\textwidth]{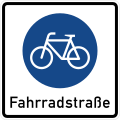} &
\textbf{Sign 244.1 –} Beginning of a cycle street. Indicates a roadway primarily designated for bicycles; motor vehicles are permitted only under specific conditions. \\
\midrule

\begin{minipage}[c]{\linewidth}
  \centering
  \includegraphics[width=0.5\textwidth]{media/icons/239.png}\\[0.3em]
  \includegraphics[width=0.5\textwidth]{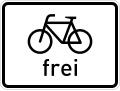}
\end{minipage}
&
\textbf{Sign 239 + Zusatzzeichen 1022-10 –} Pedestrian path with the “Cyclists allowed” supplementary plate, indicating that bicycles may also use the pedestrian walkway. \\
\midrule

\begin{minipage}[c]{\linewidth}
  \centering
  \includegraphics[width=0.5\textwidth]{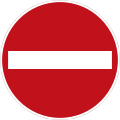}\\[0.3em]
  \includegraphics[width=0.5\textwidth]{media/icons/1022_10.png}
\end{minipage}
&
\textbf{Sign 267 + Zusatzzeichen 1022-10 –} Entry restricted with the “Cyclists allowed” supplementary plate, indicating that bicycles may enter the street. \\

\bottomrule
\end{tabular}
\caption{\textbf{Traffic signs relevant to cyclists.} Examples adapted from the Vienna Convention on Road Signs and Signals~\cite{vienna_convention_road_signs} and the German traffic regulations (\textit{Straßenverkehrs-Ordnung, StVO}).}
\label{tab:cycle_traffic_signs}
\end{table*}

\begin{table*}[!t]
	\centering

	\resizebox{\textwidth}{!}{%
		\begin{tabular}{@{} p{0.48\textwidth} p{0.52\textwidth} @{}}

			\begin{minipage}[c]{0.48\textwidth}
				\centering
                \includegraphics[width=\linewidth]{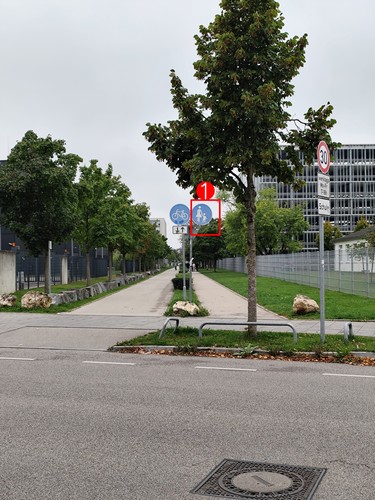}

			\end{minipage}
			 &
			\begin{minipage}[c]{0.52\textwidth}
				\begin{tcolorbox}[colback=blue!5, colframe=blue!50!black, title=\textbf{Question}, fonttitle=\scriptsize\bfseries, boxrule=0.5pt, arc=2pt, left=4pt, right=4pt, top=2pt, bottom=2pt, fontupper=\scriptsize]
					In the image, focus on the region marked as <1> (highlighted by a red bounding box). Which road sign description corresponds to the sign present in that region?
				\end{tcolorbox}

				\begin{tcolorbox}[colback=gray!5, colframe=gray!40, title=\textbf{Choices}, fonttitle=\scriptsize\bfseries, coltitle=black, boxrule=0.5pt, arc=2pt, left=4pt, right=4pt, top=2pt, bottom=2pt, fontupper=\scriptsize]
					\textbf{Options:} \\

					A. "Regulatory sign prohibiting pedestrian entry to this path" \\
					B. "Regulatory sign indicating a shared path for pedestrians and cyclists" \\
					C. "Warning sign indicating presence of children nearby, typically near schools or playgrounds" \\
					D. "Regulatory sign marking the end of a cycle-priority street" \\
					E. "Warning sign indicating narrowing of the road on the right side" \\
					F. "Regulatory sign prohibiting cycling" \\
					G. "Regulatory sign indicating a pedestrian-only path (cyclist not permitted)" \\
					\hrule \vspace{2pt}
					\textbf{{ Ground Truth:}} \textbf{G} — Regulatory sign: Pedestrian-only path (Cyclists prohibited).
				\end{tcolorbox}
			\end{minipage} \\
		\end{tabular}
	}

	\vspace{0.5em}
	\resizebox{\textwidth}{!}{%
		\begin{tabular}{@{} p{0.5\textwidth} p{0.5\textwidth} @{}}

			\begin{minipage}[t]{\linewidth}
				\begin{tcolorbox}[enhanced, colback=orange!5, colframe=orange!60!black, title=\faRobot \ \textsc{DriveLMM-O1 (8B)}, fonttitle=\small\bfseries, attach boxed title to top left={yshift=-2mm, xshift=2mm}, boxed title style={colback=orange!60!black}, boxrule=0.8pt, arc=4pt, fontupper=\scriptsize]
					\textbf{Reasoning:} The highlighted region <1> contains an icon depicting a person holding a child. This typically indicates a warning related to the presence of pedestrians, specifically near schools. \\

					\textbf{Answer:} {{C. Warning sign indicating presence of children nearby.}}
				\end{tcolorbox}
			\end{minipage}
			 &
			\begin{minipage}[t]{\linewidth}
				\begin{tcolorbox}[enhanced, colback=emerald!5, colframe=emerald!60!black, title=\faRobot \ \textsc{Qwen3-VL-8B}, fonttitle=\small\bfseries, attach boxed title to top left={yshift=-2mm, xshift=2mm}, boxed title style={colback=emerald!60!black}, boxrule=0.8pt, arc=4pt, fontupper=\scriptsize]
					\textbf{Reasoning:} The sign shows a blue circle with a white pictogram of an adult and child. The presence of a "Schule" (school) sign nearby supports the interpretation of a standard European warning sign. \\

					\textbf{Answer:} {{C. Warning sign indicating presence of children nearby.}}
				\end{tcolorbox}
			\end{minipage}
		\end{tabular}
	}

	\vspace{0.5em}
	\resizebox{\textwidth}{!}{%
		\begin{tabular}{@{} p{0.5\textwidth} p{0.5\textwidth} @{}}

			\begin{minipage}[t]{\linewidth}
				\begin{tcolorbox}[enhanced, colback=purple!5, colframe=purple!60!black, title=\faBrain \ \textsc{Cosmos-R2 (8B)}, fonttitle=\small\bfseries, attach boxed title to top left={yshift=-2mm, xshift=2mm}, boxed title style={colback=purple!60!black}, boxrule=0.8pt, arc=4pt, fontupper=\scriptsize]
					\textbf{Reasoning :} {"Blue circles usually indicate mandatory regulations rather than warnings. Option C is a warning, but the color is blue... perhaps it's a regulatory sign marking a shared path for pedestrians and cyclists, even if symbols are simplified."} \\

					\textbf{Answer:} {{B. Regulatory sign indicating a shared path for pedestrians and cyclists.}} \\
				\end{tcolorbox}
			\end{minipage}
			&

			\begin{minipage}[t]{\linewidth}
				\begin{tcolorbox}[enhanced, colback=cyan!5, colframe=cyan!60!black, title=\faRobot \ \textsc{InternVL3}, fonttitle=\small\bfseries, attach boxed title to top left={yshift=-2mm, xshift=2mm}, boxed title style={colback=cyan!60!black}, boxrule=0.8pt, arc=4pt, fontupper=\scriptsize]
					\textbf{ Reasoning:} {"Blue circles usually indicate mandatory regulations rather than warnings. Option C is a warning, but the color is blue... perhaps it's a regulatory sign marking a shared path for pedestrians and cyclists, even if symbols are simplified."} \\

					\textbf{Answer:} {{B. Regulatory sign indicating a shared path for pedestrians and cyclists.}} \\
				\end{tcolorbox}
			\end{minipage}
		\end{tabular}
	}

    \caption{\textbf{Qualitative Failure Case Analysis (TSR+S).} A regulatory semantic failure in which models correctly ground the sign but misinterpret the icon (pedestrian-only vs.\ school warning). Note that generated text is paraphrased for brevity.}
	\label{tab:casestudy1}

\end{table*}

\clearpage

\begin{table*}[!t]
	\centering

	\resizebox{\textwidth}{!}{%
		\begin{tabular}{@{} p{0.48\textwidth} p{0.52\textwidth} @{}}

			\begin{minipage}[c]{0.48\textwidth}
				\centering
                \includegraphics[width=\linewidth]{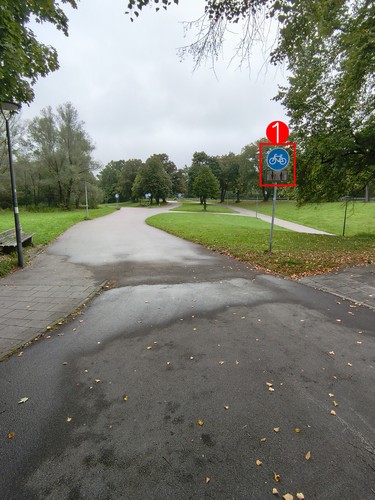}

			\end{minipage}
			 &
			\begin{minipage}[c]{0.52\textwidth}
				\begin{tcolorbox}[colback=blue!5, colframe=blue!50!black, title=\textbf{Question}, fonttitle=\small\bfseries, boxrule=0.8pt, arc=4pt, left=4pt, right=4pt, top=2pt, bottom=2pt, fontupper=\scriptsize]
					In the image, focus on the region marked as <1> (highlighted by a red bounding box). Which road sign description corresponds to the sign present in that region?
				\end{tcolorbox}

				\begin{tcolorbox}[colback=gray!5, colframe=gray!40, title=\textbf{Choices}, fonttitle=\scriptsize\bfseries, coltitle=black, boxrule=0.5pt, arc=2pt, left=4pt, right=4pt, top=2pt, bottom=2pt, fontupper=\scriptsize]
					A.  "Regulatory sign indicating a pedestrian-only path (cyclist not permitted)" \\
      B. "Combined sign indicating a parking area for the bicycles" \\
      C. "Warning sign indicating cyclists or bicycle traffic ahead"    \\
      D. "Regulatory sign indicating a shared path for pedestrians and cyclists" \\
      E. "Regulatory sign indicating parallel separated path-one for cyclists and one for pedestrians" \\
      F. "Regulatory sign prohibiting cycling" \\
      G. "Regulatory sign indicating a mandatory bicycle-only lane"

					\hrule \vspace{2pt}
					\textbf{ GT:} G — Regulatory sign indicating a mandatory bicycle-only lane
				\end{tcolorbox}
			\end{minipage} \\
		\end{tabular}
	}

	\vspace{0.5em}
	\resizebox{\textwidth}{!}{%
		\begin{tabular}{@{} p{0.5\textwidth} p{0.5\textwidth} @{}}

			\begin{minipage}[t]{\linewidth}
				\begin{tcolorbox}[enhanced, colback=orange!5, colframe=orange!60!black, title=\faRobot \ \textsc{DriveLMM-O1 (8B)}, fonttitle=\small\bfseries, attach boxed title to top left={yshift=-2mm, xshift=2mm}, boxed title style={colback=orange!60!black}, boxrule=0.8pt, arc=4pt, fontupper=\scriptsize]
					\textbf{Reasoning:} The image shows a paved path. The highlighted region shows a road sign with an icon of a bicycle and downward arrows, which typically indicate that the path is shared between cyclists and pedestrians without separation. The icons and the lack of a prohibition symbol suggest shared use. \\

					\textbf{Answer:} E. Regulatory sign indicating parallel separated paths—one for cyclists and one for pedestrians.
				\end{tcolorbox}
			\end{minipage}
			 &
			\begin{minipage}[t]{\linewidth}
				\begin{tcolorbox}[enhanced, colback=emerald!5, colframe=emerald!60!black, title=\faRobot \ \textsc{Qwen3-VL-8B}, fonttitle=\small\bfseries, attach boxed title to top left={yshift=-2mm, xshift=2mm}, boxed title style={colback=emerald!60!black}, boxrule=0.8pt, arc=4pt, fontupper=\scriptsize]
					\textbf{Reasoning:} The sign shows a blue circle with a white bicycle symbol above a rectangular sign with two arrows, which typically indicates separate paths for different users, often cyclists and pedestrians, in the same direction or in opposite directions. \\

					\textbf{Answer:} E
				\end{tcolorbox}
			\end{minipage}
		\end{tabular}
	}

	\vspace{0.5em}
	\resizebox{\textwidth}{!}{%
		\begin{tabular}{@{} p{0.5\textwidth} p{0.5\textwidth} @{}}

			\begin{minipage}[t]{\linewidth}
				\begin{tcolorbox}[enhanced, colback=purple!5, colframe=purple!60!black, title=\faBrain \ \textsc{Cosmos-R2 (8B)}, fonttitle=\small\bfseries, attach boxed title to top left={yshift=-2mm, xshift=2mm}, boxed title style={colback=purple!60!black}, boxrule=0.8pt, arc=4pt, fontupper=\scriptsize]
					\textbf{Reasoning:} The top circular blue sign with a white bicycle symbol typically means the path is designated for cyclists only. The rectangular sign below showing two arrows pointing in opposite directions usually indicates separate lanes. The combination suggests separate paths; the presence of opposing arrows implies separation rather than sharing. \\

					\textbf{Answer:} E
				\end{tcolorbox}
			\end{minipage}

			 &

			\begin{minipage}[t]{\linewidth}
				\begin{tcolorbox}[enhanced, colback=cyan!5, colframe=cyan!60!black, title=\faRobot \ \textsc{InternVL3}, fonttitle=\small\bfseries, attach boxed title to top left={yshift=-2mm, xshift=2mm}, boxed title style={colback=cyan!60!black}, boxrule=0.8pt, arc=4pt, fontupper=\scriptsize]
					\textbf{Reasoning:} The sign in the highlighted region shows a bicycle symbol with arrows indicating movement in both directions, which typically indicates a shared path for pedestrians and cyclists. \\

					\textbf{Answer:} D. Regulatory sign indicating a shared path for pedestrians and cyclists.
				\end{tcolorbox}
			\end{minipage}
		\end{tabular}
	}

	\caption{\textbf{Qualitative Failure Case Analysis (TSR+S):} Here, some models incorrectly relies on the double arrows to base its reasoning. Note that generated text is paraphrased for brevity.}
	\label{tab:casestudy2}

\end{table*}

\clearpage

\begin{table*}[!t]
	\centering

	\resizebox{\textwidth}{!}{%
		\begin{tabular}{@{} p{0.48\textwidth} p{0.52\textwidth} @{}}

			\begin{minipage}[c]{0.48\textwidth}
				\centering
                                                \includegraphics[width=\linewidth]{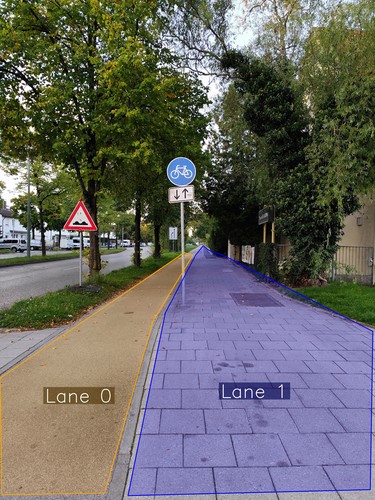}

			\end{minipage}
			 &
			\begin{minipage}[c]{0.52\textwidth}
				\begin{tcolorbox}[colback=blue!5, colframe=blue!50!black, title=\textbf{Question}, fonttitle=\small\bfseries, boxrule=0.8pt, arc=4pt, left=4pt, right=4pt, top=2pt, bottom=2pt, fontupper=\scriptsize]
					The lanes are shaded and labeled numerically. Which labeled lane corresponds to the cyclist lane?
				\end{tcolorbox}

				\begin{tcolorbox}[colback=gray!5, colframe=gray!40, title=\textbf{Choices}, fonttitle=\scriptsize\bfseries, coltitle=black, boxrule=0.5pt, arc=2pt, left=4pt, right=4pt, top=2pt, bottom=2pt, fontupper=\scriptsize]
					A. Lane 1 \\
					B. Lane 0 \\
					\hrule \vspace{2pt}
					\textbf{ GT:} B — Lane 0
				\end{tcolorbox}
			\end{minipage} \\
		\end{tabular}
	}

	\vspace{0.5em}
	\resizebox{\textwidth}{!}{%
		\begin{tabular}{@{} p{0.5\textwidth} p{0.5\textwidth} @{}}

			\begin{minipage}[t]{\linewidth}
				\begin{tcolorbox}[enhanced, colback=orange!5, colframe=orange!60!black, title=\faRobot \ \textsc{DriveLMM-O1 (8B)}, fonttitle=\small\bfseries, attach boxed title to top left={yshift=-2mm, xshift=2mm}, boxed title style={colback=orange!60!black}, boxrule=0.8pt, arc=4pt, fontupper=\scriptsize]
					\textbf{Reasoning:} Identify the relevant road signs in the image. A sign is shown indicating a bike with an arrow pointing down, which indicates a lane for bicycles. The signs are above the lanes labeled "Lane 0" and "Lane 1." Since the sign indicates "Lane 1" as having a cyclists lane, this corresponds to the intended direction of the bicycles. \\

					\textbf{Answer:} A. Lane 1
				\end{tcolorbox}
			\end{minipage}
			 &
			\begin{minipage}[t]{\linewidth}
				\begin{tcolorbox}[enhanced, colback=emerald!5, colframe=emerald!60!black, title=\faRobot \ \textsc{Qwen3-VL-8B}, fonttitle=\small\bfseries, attach boxed title to top left={yshift=-2mm, xshift=2mm}, boxed title style={colback=emerald!60!black}, boxrule=0.8pt, arc=4pt, fontupper=\scriptsize]
					\textbf{Reasoning:} The blue bicycle symbol on the sign above Lane 1 indicates it is designated for cyclists, and the lane is paved with bricks typically used for bike paths. Lane 0 is a wider, smoother surface likely for pedestrians or general traffic, not specifically for cyclists. \\

					\textbf{Answer:} A. Lane 1
				\end{tcolorbox}
			\end{minipage}
		\end{tabular}
	}

	\vspace{0.5em}
	\resizebox{\textwidth}{!}{%
		\begin{tabular}{@{} p{0.5\textwidth} p{0.5\textwidth} @{}}

			\begin{minipage}[t]{\linewidth}
				\begin{tcolorbox}[enhanced, colback=purple!5, colframe=purple!60!black, title=\faBrain \ \textsc{Cosmos-R2 (8B)}, fonttitle=\small\bfseries, attach boxed title to top left={yshift=-2mm, xshift=2mm}, boxed title style={colback=purple!60!black}, boxrule=0.8pt, arc=4pt, fontupper=\scriptsize]
					\textbf{Reasoning:} Lane 0 is a smooth, light brown asphalt path with scattered leaves. Lane 1 is made of gray rectangular paving stones. The key detail is the presence of a blue circular sign with a white bicycle symbol above Lane 1. In many countries, such signs indicate that the lane is designated for cyclists. The positioning of the sign directly over Lane 1 reinforces this association. \\

					\textbf{Answer:} A
				\end{tcolorbox}
			\end{minipage}

			 &

			\begin{minipage}[t]{\linewidth}
				\begin{tcolorbox}[enhanced, colback=cyan!5, colframe=cyan!60!black, title=\faRobot \ \textsc{InternVL3.5-8B}, fonttitle=\small\bfseries, attach boxed title to top left={yshift=-2mm, xshift=2mm}, boxed title style={colback=cyan!60!black}, boxrule=0.8pt, arc=4pt, fontupper=\scriptsize]
					\textbf{Reasoning:} The cyclist lane is indicated by the blue sign with a bicycle symbol, and it is positioned next to Lane 1, suggesting it is the designated lane for cyclists. \\

					\textbf{Answer:} A. Lane 1
				\end{tcolorbox}
			\end{minipage}
		\end{tabular}
	}

	\caption{\textbf{Failure case study for the Bicycle Lane Prediction task.} In this example, the model relies on the physical location of the signboard to infer the lane. Note that generated text is paraphrased for brevity.}
	\label{tab:casestudy3}

\end{table*}

\begin{table*}[!t]
	\centering

	\resizebox{\textwidth}{!}{%
		\begin{tabular}{@{} p{0.48\textwidth} p{0.52\textwidth} @{}}

			\begin{minipage}[c]{0.48\textwidth}
				\centering
                                \includegraphics[width=\linewidth]{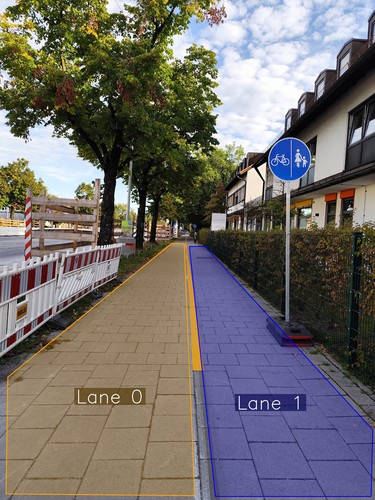}

			\end{minipage}
			 &
			\begin{minipage}[c]{0.52\textwidth}
				\begin{tcolorbox}[colback=blue!5, colframe=blue!50!black, title=\textbf{Question}, fonttitle=\small\bfseries, boxrule=0.8pt, arc=4pt, left=4pt, right=4pt, top=2pt, bottom=2pt, fontupper=\scriptsize]
					The lanes are shaded and labeled numerically. Which labeled lane corresponds to the cyclist lane?
				\end{tcolorbox}

				\begin{tcolorbox}[colback=gray!5, colframe=gray!40, title=\textbf{Choices}, fonttitle=\scriptsize\bfseries, coltitle=black, boxrule=0.5pt, arc=2pt, left=4pt, right=4pt, top=2pt, bottom=2pt, fontupper=\scriptsize]
					A. Lane 1 \\
					B. Lane 0 \\
					\hrule \vspace{2pt}
					\textbf{ GT:} B — Lane 0
				\end{tcolorbox}
			\end{minipage} \\
		\end{tabular}
	}

	\vspace{0.5em}
	\resizebox{\textwidth}{!}{%
		\begin{tabular}{@{} p{0.5\textwidth} p{0.5\textwidth} @{}}

			\begin{minipage}[t]{\linewidth}
				\begin{tcolorbox}[enhanced, colback=orange!5, colframe=orange!60!black, title=\faRobot \ \textsc{DriveLMM-O1 (8B)}, fonttitle=\small\bfseries, attach boxed title to top left={yshift=-2mm, xshift=2mm}, boxed title style={colback=orange!60!black}, boxrule=0.8pt, arc=4pt, fontupper=\scriptsize]
					\textbf{Reasoning:} In the image, there is a road sign visible that indicates the presence of a cyclist, accompanied by a bicycle symbol. The sign is located along the path, and the lane corresponding to this mark is clearly shaded as "Lane 1". Therefore, the lane intended for cycling activity is labeled as "Lane 1". \\

					\textbf{Answer:} A. Lane 1
				\end{tcolorbox}
			\end{minipage}
			 &
			\begin{minipage}[t]{\linewidth}
				\begin{tcolorbox}[enhanced, colback=emerald!5, colframe=emerald!60!black, title=\faRobot \ \textsc{Qwen3-VL-8B}, fonttitle=\small\bfseries, attach boxed title to top left={yshift=-2mm, xshift=2mm}, boxed title style={colback=emerald!60!black}, boxrule=0.8pt, arc=4pt, fontupper=\scriptsize]
					\textbf{Reasoning:} The blue circular sign with a bicycle symbol indicates a designated bike lane, and Lane 1 is shaded blue, matching the color associated with the bike lane in the image. \\

					\textbf{Answer:} A. Lane 1
				\end{tcolorbox}
			\end{minipage}
		\end{tabular}
	}

	\vspace{0.5em}
	\resizebox{\textwidth}{!}{%
		\begin{tabular}{@{} p{0.5\textwidth} p{0.5\textwidth} @{}}

			\begin{minipage}[t]{\linewidth}
				\begin{tcolorbox}[enhanced, colback=purple!5, colframe=purple!60!black, title=\faBrain \ \textsc{Cosmos-R2 (8B)}, fonttitle=\small\bfseries, attach boxed title to top left={yshift=-2mm, xshift=2mm}, boxed title style={colback=purple!60!black}, boxrule=0.8pt, arc=4pt, fontupper=\scriptsize]
					\textbf{Reasoning:} Lane 0 has light brown tiles; Lane 1 has purple tiles. The presence of a blue circular sign with a bicycle icon above Lane 1 suggests Lane 1 is designated for bicycles. Additionally, the color coding (purple for bikes) and the positioning of the sign directly adjacent to Lane 1 reinforces this association. \\

					\textbf{Answer:} A
				\end{tcolorbox}
			\end{minipage}

			 &

			\begin{minipage}[t]{\linewidth}
				\begin{tcolorbox}[enhanced, colback=cyan!5, colframe=cyan!60!black, title=\faRobot \ \textsc{InternVL3.5-8B}, fonttitle=\small\bfseries, attach boxed title to top left={yshift=-2mm, xshift=2mm}, boxed title style={colback=cyan!60!black}, boxrule=0.8pt, arc=4pt, fontupper=\scriptsize]
					\textbf{Reasoning:} Lane 1 is marked in purple with a blue line, which usually indicates a bike lane, and it also has a bicycle symbol on the sign nearby. \\

					\textbf{Answer:} A. Lane 1
				\end{tcolorbox}
			\end{minipage}
		\end{tabular}
	}

	\caption{\textbf{Failure case study for the Bicycle Lane Prediction task.} In this example, the model relies on the physical location of the signboard to infer the lane. Note that generated text is paraphrased for brevity.}
	\label{tab:casestudy4}

\end{table*}

\begin{table*}[!t]
	\centering

	\resizebox{\textwidth}{!}{%
		\begin{tabular}{@{} p{0.48\textwidth} p{0.52\textwidth} @{}}

			\begin{minipage}[c]{0.48\textwidth}
				\centering
                                \includegraphics[width=\linewidth]{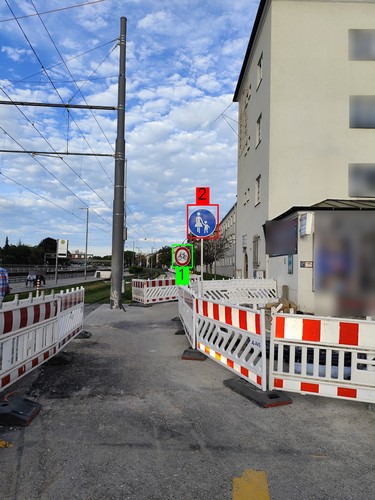}

			\end{minipage}
			 &
			\begin{minipage}[c]{0.52\textwidth}
				\begin{tcolorbox}[colback=blue!5, colframe=blue!50!black, title=\textbf{Question}, fonttitle=\small\bfseries, boxrule=0.8pt, arc=4pt, left=4pt, right=4pt, top=2pt, bottom=2pt, fontupper=\scriptsize]
					In the image, captured from the cyclist’s perspective, focus on the regions marked as <region1> (green box) and <region2> (red box). Describe the spatial relationship between the object in <region1> and the object in <region2>.
				\end{tcolorbox}

				\begin{tcolorbox}[colback=gray!5, colframe=gray!40, title=\textbf{Choices}, fonttitle=\scriptsize\bfseries, coltitle=black, boxrule=0.5pt, arc=2pt, left=4pt, right=4pt, top=2pt, bottom=2pt, fontupper=\scriptsize]
					A. The object in [region2] is closer to the cyclist than the object in [region1]. \\
					B. The object in [region1] is closer to the cyclist than the object in [region2]. \\
					\hrule \vspace{2pt}
					\textbf{ GT:} A — The object in [region2] is closer to the cyclist than the object in [region1].
				\end{tcolorbox}
			\end{minipage} \\
		\end{tabular}
	}

	\vspace{0.5em}
	\resizebox{\textwidth}{!}{%
		\begin{tabular}{@{} p{0.5\textwidth} p{0.5\textwidth} @{}}

			\begin{minipage}[t]{\linewidth}
				\begin{tcolorbox}[enhanced, colback=cyan!5, colframe=cyan!60!black, title=\faRobot \ \textsc{InternVL3.5-8B}, fonttitle=\small\bfseries, attach boxed title to top left={yshift=-2mm, xshift=2mm}, boxed title style={colback=cyan!60!black}, boxrule=0.8pt, arc=4pt, fontupper=\scriptsize]
					\textbf{Reasoning:} The green bounding box (region1) is positioned in front of the red bounding box (region2), indicating that the object in region1 is closer to the cyclist. \\

					\textbf{Answer:} B. The object in [region1] is closer to the cyclist than the object in [region2].
				\end{tcolorbox}
			\end{minipage}
			 &
			\begin{minipage}[t]{\linewidth}
				\begin{tcolorbox}[enhanced, colback=purple!5, colframe=purple!60!black, title=\faBrain \ \textsc{Cosmos-R2 (8B)}, fonttitle=\small\bfseries, attach boxed title to top left={yshift=-2mm, xshift=2mm}, boxed title style={colback=purple!60!black}, boxrule=0.8pt, arc=4pt, fontupper=\scriptsize]
					\textbf{Reasoning:} Traffic Sign 1 (left) is near the edge of the pavement, while Traffic Sign 2 (right) is mounted higher up. In the cyclist's line of sight, the left-side sign would be closer because it's on the near side of the road, whereas the higher sign might be positioned further back. \\

					\textbf{Answer:} B
				\end{tcolorbox}
			\end{minipage}
		\end{tabular}
	}

	\vspace{0.5em}
	\resizebox{\textwidth}{!}{%
		\begin{tabular}{@{} p{0.5\textwidth} p{0.5\textwidth} @{}}

			\begin{minipage}[t]{\linewidth}
				\begin{tcolorbox}[enhanced, colback=red!5, colframe=red!60!black, title=\faCube \ \textsc{SpatialThinker-7B}, fonttitle=\small\bfseries, attach boxed title to top left={yshift=-2mm, xshift=2mm}, boxed title style={colback=red!60!black}, boxrule=0.8pt, arc=4pt, fontupper=\scriptsize]
					\textbf{Reasoning:} The cyclist would see the bicycle prohibition sign (region1) first as they approach the intersection, and then the pedestrian crossing sign (region2) further ahead. The arrange shows region1 is in the foreground. \\

					\textbf{Answer:} B
				\end{tcolorbox}
			\end{minipage}

			 &

			\begin{minipage}[t]{\linewidth}
				\begin{tcolorbox}[enhanced, colback=orange!5, colframe=orange!60!black, title=\faRobot \ \textsc{VST-7B-RL}, fonttitle=\small\bfseries, attach boxed title to top left={yshift=-2mm, xshift=2mm}, boxed title style={colback=orange!60!black}, boxrule=0.8pt, arc=4pt, fontupper=\scriptsize]
					\textbf{Reasoning:} Using the concept of depth in 3D space, region1 is positioned such that it lies between the cyclist and region2. The spatial arrangement shows region1 is in the foreground relative to region2. \\

					\textbf{Answer:} B
				\end{tcolorbox}
			\end{minipage}
		\end{tabular}
	}

	\caption{\textbf{Failure case study for the Spatial Understanding (SU) task.} Here, models struggle with depth perception of the traffic elements. Note that generated text is paraphrased for brevity.}
	\label{tab:casestudy5}

\end{table*}

\begin{table*}[!t]
	\centering

	\resizebox{\textwidth}{!}{%
		\begin{tabular}{@{} p{0.48\textwidth} p{0.52\textwidth} @{}}

			\begin{minipage}[c]{0.48\textwidth}
				\centering
                \includegraphics[width=\linewidth]{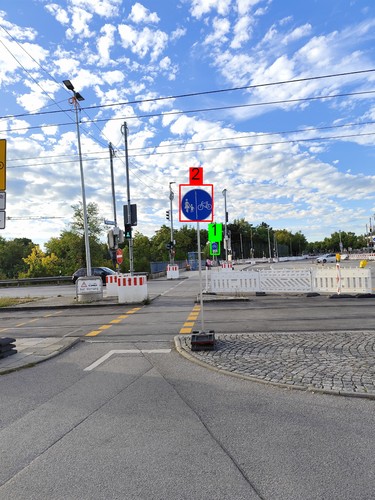}

			\end{minipage}
			 &
			\begin{minipage}[c]{0.52\textwidth}
				\begin{tcolorbox}[colback=blue!5, colframe=blue!50!black, title=\textbf{Question}, fonttitle=\small\bfseries, boxrule=0.8pt, arc=4pt, left=4pt, right=4pt, top=2pt, bottom=2pt, fontupper=\scriptsize]
					In the image, captured from the cyclist’s perspective, focus on the regions marked as <region1> (green box) and <region2> (red box). Describe the spatial relationship between the object in <region1> and the object in <region2>.
				\end{tcolorbox}

				\begin{tcolorbox}[colback=gray!5, colframe=gray!40, title=\textbf{Choices}, fonttitle=\scriptsize\bfseries, coltitle=black, boxrule=0.5pt, arc=2pt, left=4pt, right=4pt, top=2pt, bottom=2pt, fontupper=\scriptsize]
					A. The object in [region2] is closer to the cyclist than the object in [region1]. \\
					B. The object in [region1] is closer to the cyclist than the object in [region2]. \\
					\hrule \vspace{2pt}
					\textbf{\textcolor{green!50!black}{ GT:}} \textbf{A} — The object in [region2] is closer to the cyclist than the object in [region1].
				\end{tcolorbox}
			\end{minipage} \\
		\end{tabular}
	}

	\vspace{0.5em}
	\resizebox{\textwidth}{!}{%
		\begin{tabular}{@{} p{0.5\textwidth} p{0.5\textwidth} @{}}

			\begin{minipage}[t]{\linewidth}
				\begin{tcolorbox}[enhanced, colback=cyan!5, colframe=cyan!60!black, title=\faRobot \ \textsc{InternVL3.5-8B}, fonttitle=\small\bfseries, attach boxed title to top left={yshift=-2mm, xshift=2mm}, boxed title style={colback=cyan!60!black}, boxrule=0.8pt, arc=4pt, fontupper=\scriptsize]
					\textbf{Reasoning:} The cyclist's perspective shows the green sign in region1 in front of the blue sign in region2, indicating that the green sign is closer to the cyclist. \\

					\textbf{Answer:} B. The object in [region1] is closer to the cyclist than the object in [region2].
				\end{tcolorbox}
			\end{minipage}
			 &
			\begin{minipage}[t]{\linewidth}
				\begin{tcolorbox}[enhanced, colback=purple!5, colframe=purple!60!black, title=\faBrain \ \textsc{Cosmos-R2 (8B)}, fonttitle=\small\bfseries, attach boxed title to top left={yshift=-2mm, xshift=2mm}, boxed title style={colback=purple!60!black}, boxrule=0.8pt, arc=4pt, fontupper=\scriptsize]
					\textbf{Reasoning:} Objects higher on the pole appear further away because the cyclist looks upwards; objects lower down seem closer. The rectangular sign is beneath the blue one, so its position relative to the cyclist's line of sight makes it appear nearer. \\

					\textbf{Answer:} B
				\end{tcolorbox}
			\end{minipage}
		\end{tabular}
	}

	\vspace{0.5em}
	\resizebox{\textwidth}{!}{%
		\begin{tabular}{@{} p{0.5\textwidth} p{0.5\textwidth} @{}}

			\begin{minipage}[t]{\linewidth}
				\begin{tcolorbox}[enhanced, colback=red!5, colframe=red!60!black, title=\faCube \ \textsc{SpatialThinker-7B}, fonttitle=\small\bfseries, attach boxed title to top left={yshift=-2mm, xshift=2mm}, boxed title style={colback=red!60!black}, boxrule=0.8pt, arc=4pt, fontupper=\scriptsize]
					\textbf{Reasoning:} The image shows a road intersection. Region 1 is highlighted in green and is closer to the cyclist, while Region 2 is highlighted in red and is further away. Therefore, the object in Region 1 is closer. \\

					\textbf{Answer:} B
				\end{tcolorbox}
			\end{minipage}

			 &

			\begin{minipage}[t]{\linewidth}
				\begin{tcolorbox}[enhanced, colback=orange!5, colframe=orange!60!black, title=\faRobot \ \textsc{VST-7B-RL}, fonttitle=\small\bfseries, attach boxed title to top left={yshift=-2mm, xshift=2mm}, boxed title style={colback=orange!60!black}, boxrule=0.8pt, arc=4pt, fontupper=\scriptsize]
					\textbf{Reasoning:} Analyzing depth in 3D space, the object in region1 is positioned such that it lies between the cyclist and the object in region2 in terms of depth. Thus, region1 is in the foreground relative to region2. \\

					\textbf{Answer:} B
				\end{tcolorbox}
			\end{minipage}
		\end{tabular}
	}

	\caption{\textbf{Failure case study for the Spatial Understanding (SU) task.} This example highlights a systematic failure where all models incorrectly perceive depth by equating vertical height or 2D image coordinates with 3D proximity. Note that generated text is paraphrased for brevity.}
	\label{tab:casestudy6}

\end{table*}

\begin{table*}[!t]
	\centering

	\resizebox{\textwidth}{!}{%
		\begin{tabular}{@{} p{0.48\textwidth} p{0.52\textwidth} @{}}

			\begin{minipage}[c]{0.48\textwidth}
				\centering
                \includegraphics[width=\linewidth]{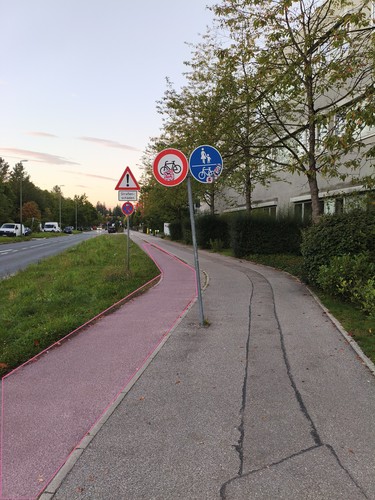}

			\end{minipage}
			 &
			\begin{minipage}[c]{0.52\textwidth}
				\begin{tcolorbox}[colback=blue!5, colframe=blue!50!black, title=\textbf{Question}, fonttitle=\small\bfseries, boxrule=0.8pt, arc=4pt, left=4pt, right=4pt, top=2pt, bottom=2pt, fontupper=\scriptsize]
					The image shows a forward-facing view from a cyclist’s perspective. From the cyclist’s perspective, does the shaded lane (in pink) ahead allow cycling? The shading is externally drawn on the image as an overlay.
				\end{tcolorbox}

				\begin{tcolorbox}[colback=gray!5, colframe=gray!40, title=\textbf{Choices}, fonttitle=\scriptsize\bfseries, coltitle=black, boxrule=0.5pt, arc=2pt, left=4pt, right=4pt, top=2pt, bottom=2pt, fontupper=\scriptsize]
					A. Yes, continue riding forward on the labelled lane \\
					B. No, cycling is not allowed on this lane \\
					\hrule \vspace{2pt}
					\textbf{\textcolor{green!50!black}{ GT:}} \textbf{B} — No, cycling is not allowed on this lane
				\end{tcolorbox}
			\end{minipage} \\
		\end{tabular}
	}

	\vspace{0.5em}
	\resizebox{\textwidth}{!}{%
		\begin{tabular}{@{} p{0.5\textwidth} p{0.5\textwidth} @{}}

			\begin{minipage}[t]{\linewidth}
				\begin{tcolorbox}[enhanced, colback=orange!5, colframe=orange!60!black, title=\faRobot \ \textsc{DriveLMM-O1 (8B)}, fonttitle=\small\bfseries, attach boxed title to top left={yshift=-2mm, xshift=2mm}, boxed title style={colback=orange!60!black}, boxrule=0.8pt, arc=4pt, fontupper=\scriptsize]
					\textbf{Reasoning:} There is a clear "Bike Route Only" sign on the far right, indicating this lane is specifically for bicycles. The pink shaded lane corresponds with the bike route as depicted by the sign. The warning triangle below it indicates caution, not prohibition. \\

					\textbf{Answer:} \textcolor{red!80!black}{\textbf{A. Yes, continue riding forward on the labelled lane}}
				\end{tcolorbox}
			\end{minipage}
			 &
			\begin{minipage}[t]{\linewidth}
				\begin{tcolorbox}[enhanced, colback=purple!5, colframe=purple!60!black, title=\faBrain \ \textsc{Cosmos-R2 (8B)}, fonttitle=\small\bfseries, attach boxed title to top left={yshift=-2mm, xshift=2mm}, boxed title style={colback=purple!60!black}, boxrule=0.8pt, arc=4pt, fontupper=\scriptsize]
					\textbf{Reasoning:} There's a red circle with a black bicycle symbol. That usually means "bicycle lane" or "cycle path," suggesting cyclists are permitted. Even if there's a shared area, the presence of the bicycle-specific sign implies cycling is allowed. \\

					\textbf{Answer:} \textcolor{red!80!black}{\textbf{A}}
				\end{tcolorbox}
			\end{minipage}
		\end{tabular}
	}

	\vspace{0.5em}
	\resizebox{\textwidth}{!}{%
		\begin{tabular}{@{} p{0.5\textwidth} p{0.5\textwidth} @{}}

			\begin{minipage}[t]{\linewidth}
				\begin{tcolorbox}[enhanced, colback=emerald!5, colframe=emerald!60!black, title=\faRobot \ \textsc{Qwen3-VL-8B}, fonttitle=\small\bfseries, attach boxed title to top left={yshift=-2mm, xshift=2mm}, boxed title style={colback=emerald!60!black}, boxrule=0.8pt, arc=4pt, fontupper=\scriptsize]
					\textbf{Reasoning:} The red circle with a bicycle symbol indicates a "No Cycling" zone. Therefore, cycling is not permitted on the shaded lane. \\
					\hrule \vspace{2pt}
					\textbf{Answer:} \textcolor{green!50!black}{\textbf{B. No, cycling is not allowed on this lane}}
				\end{tcolorbox}
			\end{minipage}

			 &

			\begin{minipage}[t]{\linewidth}
				\begin{tcolorbox}[enhanced, colback=cyan!5, colframe=cyan!60!black, title=\faRobot \ \textsc{InternVL3.5-8B}, fonttitle=\small\bfseries, attach boxed title to top left={yshift=-2mm, xshift=2mm}, boxed title style={colback=cyan!60!black}, boxrule=0.8pt, arc=4pt, fontupper=\scriptsize]
					\textbf{Reasoning:} The traffic sign shows a symbol of a bicycle with a red circle, which means that cycling is prohibited in this lane. The lane also has a "Straßen schaden" (road damage) warning. \\
					\hrule \vspace{2pt}
					\textbf{Answer:} \textcolor{green!50!black}{\textbf{B. No, cycling is not allowed on this lane}}
				\end{tcolorbox}
			\end{minipage}
		\end{tabular}
	}

	\caption{\textbf{Failure case study for the Sign-Action Association (SAA) task.} This example demonstrates the lack of traffic rule understanding error, where some models correctly identify the bicycle pictogram but misinterpret the red circular border as a permissive indicator rather than a prohibition. Note that generated text is paraphrased for brevity.}
	\label{tab:casestudy7}

\end{table*}

\begin{table*}[!t]
	\centering

	\resizebox{\textwidth}{!}{%
		\begin{tabular}{@{} p{0.48\textwidth} p{0.52\textwidth} @{}}

			\begin{minipage}[c]{0.48\textwidth}
				\centering
                \includegraphics[width=\linewidth]{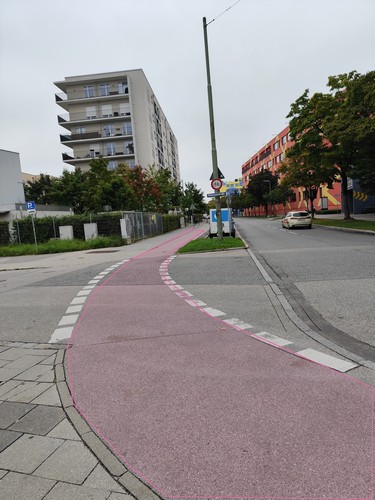}

			\end{minipage}
			 &
			\begin{minipage}[c]{0.52\textwidth}
				\begin{tcolorbox}[colback=blue!5, colframe=blue!50!black, title=\textbf{Question}, fonttitle=\small\bfseries, boxrule=0.8pt, arc=4pt, left=4pt, right=4pt, top=2pt, bottom=2pt, fontupper=\scriptsize]
					The image shows a forward-facing view from a cyclist’s perspective. From the cyclist’s forward view, is cycling allowed along the highlighted lane (in pink)? The shading is externally drawn on the image as an overlay.
				\end{tcolorbox}

				\begin{tcolorbox}[colback=gray!5, colframe=gray!40, title=\textbf{Choices}, fonttitle=\scriptsize\bfseries, coltitle=black, boxrule=0.5pt, arc=2pt, left=4pt, right=4pt, top=2pt, bottom=2pt, fontupper=\scriptsize]
					A. No, cycling is not allowed on this lane \\
					B. Yes, continue riding forward on the labelled lane \\
					\hrule \vspace{2pt}
					\textbf{\textcolor{green!50!black}{ GT:}} \textbf{A} — No, cycling is not allowed on this lane
				\end{tcolorbox}
			\end{minipage} \\
		\end{tabular}
	}

	\vspace{0.5em}
	\resizebox{\textwidth}{!}{%
		\begin{tabular}{@{} p{0.5\textwidth} p{0.5\textwidth} @{}}

			\begin{minipage}[t]{\linewidth}
				\begin{tcolorbox}[enhanced, colback=orange!5, colframe=orange!60!black, title=\faRobot \ \textsc{DriveLMM-O1 (8B)}, fonttitle=\small\bfseries, attach boxed title to top left={yshift=-2mm, xshift=2mm}, boxed title style={colback=orange!60!black}, boxrule=0.8pt, arc=4pt, fontupper=\scriptsize]
					\textbf{Reasoning:} The lane is pink, indicating a designated cycling path. There is a "P" sign visible for parking, but no specific prohibition shown. The lack of obstructions or signs prohibiting biking directly implies that it is likely permitted. \\

					\textbf{Answer:} \textcolor{red!80!black}{\textbf{B. Yes, continue riding forward on the labelled lane.}}
				\end{tcolorbox}
			\end{minipage}
			 &
			\begin{minipage}[t]{\linewidth}
				\begin{tcolorbox}[enhanced, colback=purple!5, colframe=purple!60!black, title=\faBrain \ \textsc{Cosmos-R2 (8B)}, fonttitle=\small\bfseries, attach boxed title to top left={yshift=-2mm, xshift=2mm}, boxed title style={colback=purple!60!black}, boxrule=0.8pt, arc=4pt, fontupper=\scriptsize]
					\textbf{Reasoning:} The scene includes a bicycle lane marked by dashed lines and a red bike symbol on a pole. Such symbols indicate designated paths. Since the lane is specifically marked for bicycles, it suggests cycling is permitted. \\

					\textbf{Answer:} \textcolor{red!80!black}{\textbf{B}}
				\end{tcolorbox}
			\end{minipage}
		\end{tabular}
	}

	\vspace{0.5em}
	\resizebox{\textwidth}{!}{%
		\begin{tabular}{@{} p{0.5\textwidth} p{0.5\textwidth} @{}}

			\begin{minipage}[t]{\linewidth}
				\begin{tcolorbox}[enhanced, colback=emerald!5, colframe=emerald!60!black, title=\faRobot \ \textsc{Qwen3-VL-8B}, fonttitle=\small\bfseries, attach boxed title to top left={yshift=-2mm, xshift=2mm}, boxed title style={colback=emerald!60!black}, boxrule=0.8pt, arc=4pt, fontupper=\scriptsize]
					\textbf{Reasoning:} The bicycle symbol on the traffic sign (red circle, white background, black icon) typically indicates that cycling is permitted and encouraged. The lane is physically separated by a grass strip, confirming it is a bike lane. \\

					\textbf{Answer:} \textcolor{red!80!black}{\textbf{B. Yes, continue riding forward on the labelled lane}}
				\end{tcolorbox}
			\end{minipage}

			 &

			\begin{minipage}[t]{\linewidth}
				\begin{tcolorbox}[enhanced, colback=cyan!5, colframe=cyan!60!black, title=\faRobot \ \textsc{InternVL3.5-8B}, fonttitle=\small\bfseries, attach boxed title to top left={yshift=-2mm, xshift=2mm}, boxed title style={colback=cyan!60!black}, boxrule=0.8pt, arc=4pt, fontupper=\scriptsize]
					\textbf{Reasoning:} The highlighted pink lane is a designated bicycle path, evident from the sign showing a bicycle symbol. The forward-facing view from the cyclist's perspective shows they are in the correct lane. \\

					\textbf{Answer:} \textcolor{red!80!black}{\textbf{B. Yes, continue riding forward on the labelled lane}}
				\end{tcolorbox}
			\end{minipage}
		\end{tabular}
	}
	\caption{\textbf{Failure case study for the Sign-Action Association (SAA) task.} This case illustrates a high-frequency failure where all models misinterpret a red circular prohibitory sign as a permissive one. Note that generated text is paraphrased for brevity.}
	\label{tab:casestudy8}

\end{table*}

\begin{table*}[!t]
	\centering

	\resizebox{\textwidth}{!}{%
		\begin{tabular}{@{} p{0.48\textwidth} p{0.52\textwidth} @{}}

			\begin{minipage}[c]{0.48\textwidth}
				\centering
                \includegraphics[width=\linewidth]{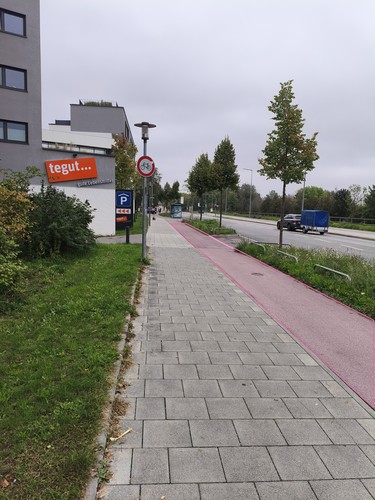}

			\end{minipage}
			 &
			\begin{minipage}[c]{0.52\textwidth}
				\begin{tcolorbox}[colback=blue!5, colframe=blue!50!black, title=\textbf{Question}, fonttitle=\small\bfseries, boxrule=0.8pt, arc=4pt, left=4pt, right=4pt, top=2pt, bottom=2pt, fontupper=\scriptsize]
					The image shows a forward-facing view from a cyclist’s perspective. From the cyclist’s forward view, is cycling allowed along the highlighted lane (in pink)? The shading is externally drawn on the image as an overlay.
				\end{tcolorbox}

				\begin{tcolorbox}[colback=gray!5, colframe=gray!40, title=\textbf{Choices}, fonttitle=\scriptsize\bfseries, coltitle=black, boxrule=0.5pt, arc=2pt, left=4pt, right=4pt, top=2pt, bottom=2pt, fontupper=\scriptsize]
					A. No, cycling is not allowed on this lane \\
					B. Yes, continue riding forward on the labelled lane \\
					\hrule \vspace{2pt}
					\textbf{\textcolor{green!50!black}{ GT:}} \textbf{A} — No, cycling is not allowed on this lane
				\end{tcolorbox}
			\end{minipage} \\
		\end{tabular}
	}

	\vspace{0.5em}
	\resizebox{\textwidth}{!}{%
		\begin{tabular}{@{} p{0.5\textwidth} p{0.5\textwidth} @{}}

			\begin{minipage}[t]{\linewidth}
				\begin{tcolorbox}[enhanced, colback=orange!5, colframe=orange!60!black, title=\faRobot \ \textsc{DriveLMM-O1 (8B)}, fonttitle=\small\bfseries, attach boxed title to top left={yshift=-2mm, xshift=2mm}, boxed title style={colback=orange!60!black}, boxrule=0.8pt, arc=4pt, fontupper=\scriptsize]
					\textbf{Reasoning:} Pay close attention to the highlighted lane in pink. Notice the sign on the left indicating parking. The presence of this sign, alongside the specific regulatory markings, suggest restricted use where cycling is not allowed. \\
					\hrule \vspace{2pt}
					\textbf{Answer:} \textcolor{green!50!black}{\textbf{A. No, cycling is not allowed on this lane.}}
				\end{tcolorbox}
			\end{minipage}
			 &
			\begin{minipage}[t]{\linewidth}
				\begin{tcolorbox}[enhanced, colback=purple!5, colframe=purple!60!black, title=\faBrain \ \textsc{Cosmos-R2 (8B)}, fonttitle=\small\bfseries, attach boxed title to top left={yshift=-2mm, xshift=2mm}, boxed title style={colback=purple!60!black}, boxrule=0.8pt, arc=4pt, fontupper=\scriptsize]
					\textbf{Reasoning:} There's a red circular sign with a black bicycle symbol. That usually indicates that cycling is permitted there. Since the sign explicitly marks the lane for bicycles, the correct answer should be that the cyclist can continue using it. \\

					\textbf{Answer:} \textcolor{red!80!black}{\textbf{B}}
				\end{tcolorbox}
			\end{minipage}
		\end{tabular}
	}

	\vspace{0.5em}
	\resizebox{\textwidth}{!}{%
		\begin{tabular}{@{} p{0.5\textwidth} p{0.5\textwidth} @{}}

			\begin{minipage}[t]{\linewidth}
				\begin{tcolorbox}[enhanced, colback=emerald!5, colframe=emerald!60!black, title=\faRobot \ \textsc{Qwen3-VL-8B}, fonttitle=\small\bfseries, attach boxed title to top left={yshift=-2mm, xshift=2mm}, boxed title style={colback=emerald!60!black}, boxrule=0.8pt, arc=4pt, fontupper=\scriptsize]
					\textbf{Reasoning:} The pink highlighted lane is a designated bike path. The signpost shows a bicycle symbol within a red circle, which typically indicates a bike route. Since the cyclist is on the path and the sign suggests it is for cyclists, cycling is allowed. \\

					\textbf{Answer:} \textcolor{red!80!black}{\textbf{B. Yes, continue riding forward on the labelled lane}}
				\end{tcolorbox}
			\end{minipage}

			 &

			\begin{minipage}[t]{\linewidth}
				\begin{tcolorbox}[enhanced, colback=cyan!5, colframe=cyan!60!black, title=\faRobot \ \textsc{InternVL3.5-8B}, fonttitle=\small\bfseries, attach boxed title to top left={yshift=-2mm, xshift=2mm}, boxed title style={colback=cyan!60!black}, boxrule=0.8pt, arc=4pt, fontupper=\scriptsize]
					\textbf{Reasoning:} The image shows a no-bicycle sign above the parking sign, indicating that cycling is not permitted on this lane. The red circular border serves as the primary prohibitory indicator. \\
					\hrule \vspace{2pt}
					\textbf{Answer:} \textcolor{green!50!black}{\textbf{A. No, cycling is not allowed on this lane}}
				\end{tcolorbox}
			\end{minipage}
		\end{tabular}
	}

	\caption{\textbf{Failure case study for the Sign-Action Association (SAA) task.} This example traffic rule misunderstanding error. In particular, models correctly identify the bicycle icon and the red circle but fail to associate the red circular frame with a prohibition, misinterpreting it as a permissive marker. Note that generated text is paraphrased for brevity.}
	\label{tab:casestudy9}

\end{table*}

\begin{table*}[!t]
	\centering

	\resizebox{\textwidth}{!}{%
		\begin{tabular}{@{} p{0.48\textwidth} p{0.52\textwidth} @{}}

			\begin{minipage}[c]{0.48\textwidth}
				\centering

                \includegraphics[width=\linewidth]{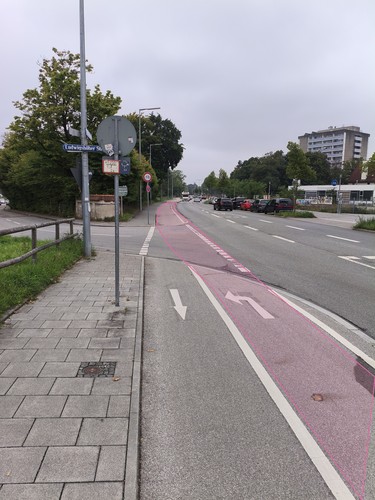}

			\end{minipage}
			 &
			\begin{minipage}[c]{0.52\textwidth}
				\begin{tcolorbox}[colback=blue!5, colframe=blue!50!black, title=\textbf{Question}, fonttitle=\small\bfseries, boxrule=0.8pt, arc=4pt, left=4pt, right=4pt, top=2pt, bottom=2pt, fontupper=\scriptsize]
					The image shows a forward-facing view from a cyclist’s perspective. Considering the shaded lane (in pink) ahead, should the cyclist stay on this route or avoid it? The shading is externally drawn on the image as an overlay.
				\end{tcolorbox}

				\begin{tcolorbox}[colback=gray!5, colframe=gray!40, title=\textbf{Choices}, fonttitle=\scriptsize\bfseries, coltitle=black, boxrule=0.5pt, arc=2pt, left=4pt, right=4pt, top=2pt, bottom=2pt, fontupper=\scriptsize]
					A. No, cycling is not allowed on this lane \\
					B. Yes, continue riding forward on the labelled lane \\
					\hrule \vspace{2pt}
					\textbf{\textcolor{green!50!black}{ GT:}} \textbf{A} — No, cycling is not allowed on this lane
				\end{tcolorbox}
			\end{minipage} \\
		\end{tabular}
	}

	\vspace{0.5em}
	\resizebox{\textwidth}{!}{%
		\begin{tabular}{@{} p{0.5\textwidth} p{0.5\textwidth} @{}}

			\begin{minipage}[t]{\linewidth}
				\begin{tcolorbox}[enhanced, colback=orange!5, colframe=orange!60!black, title=\faRobot \ \textsc{DriveLMM-O1 (8B)}, fonttitle=\small\bfseries, attach boxed title to top left={yshift=-2mm, xshift=2mm}, boxed title style={colback=orange!60!black}, boxrule=0.8pt, arc=4pt, fontupper=\scriptsize]
					\textbf{Reasoning:} The image shows a lane designated for cyclists by the arrow on the road. The pink shaded area indicates this lane. There is a "no entry" sign for vehicles nearby, indicating this area is reserved for cycling. \\

					\textbf{Answer:} \textcolor{red!80!black}{\textbf{B. Yes, continue riding forward on the labelled lane.}}
				\end{tcolorbox}
			\end{minipage}
			 &
			\begin{minipage}[t]{\linewidth}
				\begin{tcolorbox}[enhanced, colback=purple!5, colframe=purple!60!black, title=\faBrain \ \textsc{Cosmos-R2 (8B)}, fonttitle=\small\bfseries, attach boxed title to top left={yshift=-2mm, xshift=2mm}, boxed title style={colback=purple!60!black}, boxrule=0.8pt, arc=4pt, fontupper=\scriptsize]
					\textbf{Reasoning:} There's a bicycle lane marked by arrows pointing left. The pink area is part of that lane. The presence of the bike symbol suggests it's allowed. Continuing on that lane would be appropriate unless obstacles exist. \\

					\textbf{Answer:} \textcolor{red!80!black}{\textbf{B}}
				\end{tcolorbox}
			\end{minipage}
		\end{tabular}
	}

	\vspace{0.5em}
	\resizebox{\textwidth}{!}{%
		\begin{tabular}{@{} p{0.5\textwidth} p{0.5\textwidth} @{}}

			\begin{minipage}[t]{\linewidth}
				\begin{tcolorbox}[enhanced, colback=emerald!5, colframe=emerald!60!black, title=\faRobot \ \textsc{Qwen3-VL-8B}, fonttitle=\small\bfseries, attach boxed title to top left={yshift=-2mm, xshift=2mm}, boxed title style={colback=emerald!60!black}, boxrule=0.8pt, arc=4pt, fontupper=\scriptsize]
					\textbf{Reasoning:} The visual cues, including the lane markings and the bicycle pictogram on the sign, suggest a designated route for cyclists. Forward movement on the pink shaded lane is therefore permitted. \\

					\textbf{Answer:} \textcolor{red!80!black}{\textbf{B. Yes, continue riding forward on the labelled lane}}
				\end{tcolorbox}
			\end{minipage}

			 &

			\begin{minipage}[t]{\linewidth}
				\begin{tcolorbox}[enhanced, colback=cyan!5, colframe=cyan!60!black, title=\faRobot \ \textsc{InternVL3.5-8B}, fonttitle=\small\bfseries, attach boxed title to top left={yshift=-2mm, xshift=2mm}, boxed title style={colback=cyan!60!black}, boxrule=0.8pt, arc=4pt, fontupper=\scriptsize]
					\textbf{Reasoning:} The presence of a no-entry sign for cyclists (red circle with a white bike inside) indicates that cycling is prohibited in this lane. This regulatory sign overrides other visual indicators. \\
					\hrule \vspace{2pt}
					\textbf{Answer:} \textcolor{green!50!black}{\textbf{A. No, cycling is not allowed on this lane}}
				\end{tcolorbox}
			\end{minipage}
		\end{tabular}
	}
	\caption{\textbf{Failure case study for the Sign-Action Association (SAA) task.} This case highlights a critical failure in directional and regulatory sign interpretation. Most models misinterpreted the lane markings and the prohibitory sign, incorrectly concluding that the path was permissive for forward travel. Note that generated text is paraphrased for brevity.}
	\label{tab:casestudy10}

\end{table*}

\begin{table*}[!t]
	\centering

	\resizebox{\textwidth}{!}{%
		\begin{tabular}{@{} p{0.48\textwidth} p{0.52\textwidth} @{}}

			\begin{minipage}[c]{0.48\textwidth}
				\centering

                \includegraphics[width=\linewidth]{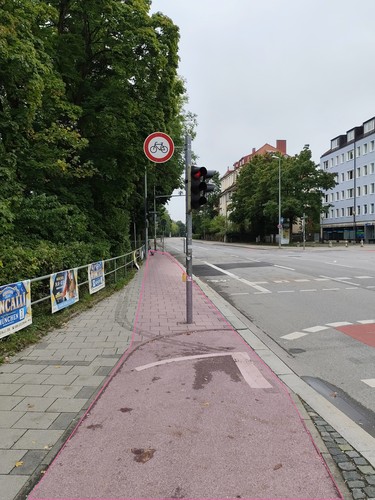}

			\end{minipage}
			 &
			\begin{minipage}[c]{0.52\textwidth}
				\begin{tcolorbox}[colback=blue!5, colframe=blue!50!black, title=\textbf{Question}, fonttitle=\small\bfseries, boxrule=0.8pt, arc=4pt, left=4pt, right=4pt, top=2pt, bottom=2pt, fontupper=\scriptsize]
					The image shows a forward-facing view from a cyclist’s perspective. From the cyclist’s perspective, does the shaded lane (in pink) ahead allow cycling? The shading is externally drawn on the image as an overlay.
				\end{tcolorbox}

				\begin{tcolorbox}[colback=gray!5, colframe=gray!40, title=\textbf{Choices}, fonttitle=\scriptsize\bfseries, coltitle=black, boxrule=0.5pt, arc=2pt, left=4pt, right=4pt, top=2pt, bottom=2pt, fontupper=\scriptsize]
					A. Yes, continue riding forward on the labelled lane \\
					B. No, cycling is not allowed on this lane \\
					\hrule \vspace{2pt}
					\textbf{\textcolor{green!50!black}{ GT:}} \textbf{B} — No, cycling is not allowed on this lane
				\end{tcolorbox}
			\end{minipage} \\
		\end{tabular}
	}

	\vspace{0.5em}
	\resizebox{\textwidth}{!}{%
		\begin{tabular}{@{} p{0.5\textwidth} p{0.5\textwidth} @{}}

			\begin{minipage}[t]{\linewidth}
				\begin{tcolorbox}[enhanced, colback=orange!5, colframe=orange!60!black, title=\faRobot \ \textsc{DriveLMM-O1 (8B)}, fonttitle=\small\bfseries, attach boxed title to top left={yshift=-2mm, xshift=2mm}, boxed title style={colback=orange!60!black}, boxrule=0.8pt, arc=4pt, fontupper=\scriptsize]
					\textbf{Reasoning:} The image shows a sign indicating a bicycle lane. Currently, the light is green indicating it is safe to proceed. The lane is explicitly colored and separated from the road, further confirming its intent for bikers. \\

					\textbf{Answer:} \textcolor{red!80!black}{\textbf{A. Yes, continue riding forward on the labelled lane.}}
				\end{tcolorbox}
			\end{minipage}
			 &
			\begin{minipage}[t]{\linewidth}
				\begin{tcolorbox}[enhanced, colback=purple!5, colframe=purple!60!black, title=\faBrain \ \textsc{Cosmos-R2 (8B)}, fonttitle=\small\bfseries, attach boxed title to top left={yshift=-2mm, xshift=2mm}, boxed title style={colback=purple!60!black}, boxrule=0.8pt, arc=4pt, fontupper=\scriptsize]
					\textbf{Reasoning:} The paved area specifically mentions it's designated for cyclists. Since it's labeled as a cyclist's paved area, that suggests it's meant for cycling. You can continue riding forward on that lane. \\

					\textbf{Answer:} \textcolor{red!80!black}{\textbf{A}}
				\end{tcolorbox}
			\end{minipage}
		\end{tabular}
	}

	\vspace{0.5em}
	\resizebox{\textwidth}{!}{%
		\begin{tabular}{@{} p{0.5\textwidth} p{0.5\textwidth} @{}}

			\begin{minipage}[t]{\linewidth}
				\begin{tcolorbox}[enhanced, colback=emerald!5, colframe=emerald!60!black, title=\faRobot \ \textsc{Qwen3-VL-8B}, fonttitle=\small\bfseries, attach boxed title to top left={yshift=-2mm, xshift=2mm}, boxed title style={colback=emerald!60!black}, boxrule=0.8pt, arc=4pt, fontupper=\scriptsize]
					\textbf{Reasoning:} The red circle with a bicycle symbol indicates a "No Bicycles" zone, meaning cycling is prohibited in this area regardless of other lane markings. \\
					\hrule \vspace{2pt}
					\textbf{Answer:} \textcolor{green!50!black}{\textbf{B. No, cycling is not allowed on this lane}}
				\end{tcolorbox}
			\end{minipage}

			 &

			\begin{minipage}[t]{\linewidth}
				\begin{tcolorbox}[enhanced, colback=cyan!5, colframe=cyan!60!black, title=\faRobot \ \textsc{InternVL3.5-8B}, fonttitle=\small\bfseries, attach boxed title to top left={yshift=-2mm, xshift=2mm}, boxed title style={colback=cyan!60!black}, boxrule=0.8pt, arc=4pt, fontupper=\scriptsize]
					\textbf{Reasoning:} The lane is marked with a traffic sign indicating it is a no-bicycle zone. Prohibitory signs take precedence over the existence of a paved path. \\
					\hrule \vspace{2pt}
					\textbf{Answer:} \textcolor{green!50!black}{\textbf{B. No, cycling is not allowed on this lane}}
				\end{tcolorbox}
			\end{minipage}
		\end{tabular}
	}

	\caption{\textbf{Failure case study for the Sign-Action Association (SAA) task.} This example highlights a traffic rule misunderstanding error where some models incorrectly rely on adjacent traffic lights while ignoring relevant regulatory "No Bicycles" sign. Note that generated text is paraphrased for brevity.}
	\label{tab:casestudy11}

\end{table*}

\begin{table*}[!t]
	\centering

	\resizebox{\textwidth}{!}{%
		\begin{tabular}{@{} p{0.48\textwidth} p{0.52\textwidth} @{}}

			\begin{minipage}[c]{0.48\textwidth}
				\centering
                \includegraphics[width=\linewidth]{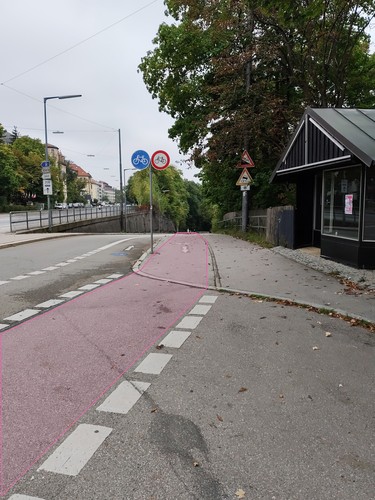}

			\end{minipage}
			 &
			\begin{minipage}[c]{0.52\textwidth}
				\begin{tcolorbox}[colback=blue!5, colframe=blue!50!black, title=\textbf{Question}, fonttitle=\small\bfseries, boxrule=0.8pt, arc=4pt, left=4pt, right=4pt, top=2pt, bottom=2pt, fontupper=\scriptsize]
					The image shows a forward-facing view from a cyclist’s perspective. From the cyclist’s forward view, is cycling allowed along the highlighted lane (in pink)? The shading is externally drawn on the image as an overlay.
				\end{tcolorbox}

				\begin{tcolorbox}[colback=gray!5, colframe=gray!40, title=\textbf{Choices}, fonttitle=\scriptsize\bfseries, coltitle=black, boxrule=0.5pt, arc=2pt, left=4pt, right=4pt, top=2pt, bottom=2pt, fontupper=\scriptsize]
					A. Yes, continue riding forward on the labelled lane \\
					B. No, cycling is not allowed on this lane \\
					\hrule \vspace{2pt}
					\textbf{\textcolor{green!50!black}{ GT:}} \textbf{B} — No, cycling is not allowed on this lane
				\end{tcolorbox}
			\end{minipage} \\
		\end{tabular}
	}

	\vspace{0.5em}
	\resizebox{\textwidth}{!}{%
		\begin{tabular}{@{} p{0.5\textwidth} p{0.5\textwidth} @{}}

			\begin{minipage}[t]{\linewidth}
				\begin{tcolorbox}[enhanced, colback=orange!5, colframe=orange!60!black, title=\faRobot \ \textsc{DriveLMM-O1 (8B)}, fonttitle=\small\bfseries, attach boxed title to top left={yshift=-2mm, xshift=2mm}, boxed title style={colback=orange!60!black}, boxrule=0.8pt, arc=4pt, fontupper=\scriptsize]
					\textbf{Reasoning:} The pink lane suggests it is specifically designated for cyclists. The arrangement of markings and indicators confirms that this is a lane intended for cyclists and indicates that it is safe to proceed. \\

					\textbf{Answer:} \textcolor{red!80!black}{\textbf{A. Yes, continue riding forward on the labelled lane.}}
				\end{tcolorbox}
			\end{minipage}
			 &
			\begin{minipage}[t]{\linewidth}
				\begin{tcolorbox}[enhanced, colback=purple!5, colframe=purple!60!black, title=\faBrain \ \textsc{Cosmos-R2 (8B)}, fonttitle=\small\bfseries, attach boxed title to top left={yshift=-2mm, xshift=2mm}, boxed title style={colback=purple!60!black}, boxrule=0.8pt, arc=4pt, fontupper=\scriptsize]
					\textbf{Reasoning:} There's a pink lane with white arrows and a blue circular bicycle sign. Even though there's a red circle nearby, the blue sign over the lane overrides it. The primary indicator for the highlighted lane allows cycling. \\

					\textbf{Answer:} \textcolor{red!80!black}{\textbf{A}}
				\end{tcolorbox}
			\end{minipage}
		\end{tabular}
	}

	\vspace{0.5em}
	\resizebox{\textwidth}{!}{%
		\begin{tabular}{@{} p{0.5\textwidth} p{0.5\textwidth} @{}}

			\begin{minipage}[t]{\linewidth}
				\begin{tcolorbox}[enhanced, colback=emerald!5, colframe=emerald!60!black, title=\faRobot \ \textsc{Qwen3-VL-8B}, fonttitle=\small\bfseries, attach boxed title to top left={yshift=-2mm, xshift=2mm}, boxed title style={colback=emerald!60!black}, boxrule=0.8pt, arc=4pt, fontupper=\scriptsize]
					\textbf{Reasoning:} The blue circular sign indicates permitted cycling. The "no cycling" sign is positioned above the pink lane, suggesting it applies to the area beyond the section, not the highlighted lane itself. The pink color confirms it is allowed. \\

					\textbf{Answer:} \textcolor{red!80!black}{\textbf{A. Yes, continue riding forward on the labelled lane}}
				\end{tcolorbox}
			\end{minipage}

			 &

			\begin{minipage}[t]{\linewidth}
				\begin{tcolorbox}[enhanced, colback=cyan!5, colframe=cyan!60!black, title=\faRobot \ \textsc{InternVL3.5-8B}, fonttitle=\small\bfseries, attach boxed title to top left={yshift=-2mm, xshift=2mm}, boxed title style={colback=cyan!60!black}, boxrule=0.8pt, arc=4pt, fontupper=\scriptsize]
					\textbf{Reasoning:} The signs indicate that cycling is prohibited in this lane due to the red circle with a line through the bicycle symbol. This prohibitory status remains the primary regulatory instruction. \\
					\hrule \vspace{2pt}
					\textbf{Answer:} \textcolor{green!50!black}{\textbf{B. No, cycling is not allowed on this lane}}
				\end{tcolorbox}
			\end{minipage}
		\end{tabular}
	}

	\caption{\textbf{Failure case study for the Sign-Action Association (SAA) task.} This case highlights a Lane-misassociation error where models struggle to resolve contradictory signals (prohibitory vs. permissive) and fail to assign to correct lanes. Note that generated text is paraphrased for brevity.
	}
	\label{tab:casestudy12}
\end{table*}

\begin{table*}[!t]
	\centering

	\resizebox{\textwidth}{!}{%
		\begin{tabular}{@{} p{0.48\textwidth} p{0.52\textwidth} @{}}

			\begin{minipage}[c]{0.48\textwidth}
				\centering

                \includegraphics[width=\linewidth]{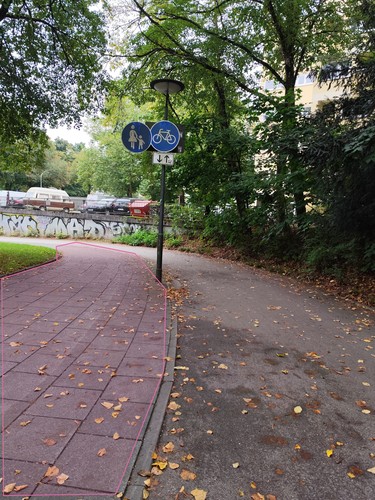}

			\end{minipage}
			 &
			\begin{minipage}[c]{0.52\textwidth}
				\begin{tcolorbox}[colback=blue!5, colframe=blue!50!black, title=\textbf{Question}, fonttitle=\small\bfseries, boxrule=0.8pt, arc=4pt, left=4pt, right=4pt, top=2pt, bottom=2pt, fontupper=\scriptsize]
					The image shows a forward-facing view from a cyclist’s perspective. Looking at the shaded lane (in pink) in the image, is the cyclist allowed to keep following it? The shading is externally drawn on the image as an overlay.
				\end{tcolorbox}

				\begin{tcolorbox}[colback=gray!5, colframe=gray!40, title=\textbf{Choices}, fonttitle=\scriptsize\bfseries, coltitle=black, boxrule=0.5pt, arc=2pt, left=4pt, right=4pt, top=2pt, bottom=2pt, fontupper=\scriptsize]
					A. Yes, continue riding forward on the shaded lane \\
					B. No, cycling is not allowed on this lane \\
					\hrule \vspace{2pt}
					\textbf{\textcolor{green!50!black}{ GT:}} \textbf{B} — No, cycling is not allowed on this lane
				\end{tcolorbox}
			\end{minipage} \\
		\end{tabular}
	}

	\vspace{0.5em}
	\resizebox{\textwidth}{!}{%
		\begin{tabular}{@{} p{0.5\textwidth} p{0.5\textwidth} @{}}

			\begin{minipage}[t]{\linewidth}
				\begin{tcolorbox}[enhanced, colback=orange!5, colframe=orange!60!black, title=\faRobot \ \textsc{DriveLMM-O1 (8B)}, fonttitle=\small\bfseries, attach boxed title to top left={yshift=-2mm, xshift=2mm}, boxed title style={colback=orange!60!black}, boxrule=0.8pt, arc=4pt, fontupper=\scriptsize]
					\textbf{Reasoning:} The blue circular sign with a bicycle symbol indicates this is a bike lane. The pink shaded lane corresponds with the bike lane marked by the sign. Since the lane is marked for cycle use, the cyclist is permitted to continue. \\

					\textbf{Answer:} \textcolor{red!80!black}{\textbf{A. Yes, continue riding forward on the shaded lane}}
				\end{tcolorbox}
			\end{minipage}
			 &
			\begin{minipage}[t]{\linewidth}
				\begin{tcolorbox}[enhanced, colback=purple!5, colframe=purple!60!black, title=\faBrain \ \textsc{Cosmos-R2 (8B)}, fonttitle=\small\bfseries, attach boxed title to top left={yshift=-2mm, xshift=2mm}, boxed title style={colback=purple!60!black}, boxrule=0.8pt, arc=4pt, fontupper=\scriptsize]
					\textbf{Reasoning:} There's a blue circular sign with a bicycle symbol and another with pedestrians. Usually, these shared spaces allow cyclists to proceed unless a red prohibition sign is present. The shaded lane aligns with the bicycle indicator. \\

					\textbf{Answer:} \textcolor{red!80!black}{\textbf{A}}
				\end{tcolorbox}
			\end{minipage}
		\end{tabular}
	}

	\vspace{0.5em}
	\resizebox{\textwidth}{!}{%
		\begin{tabular}{@{} p{0.5\textwidth} p{0.5\textwidth} @{}}

			\begin{minipage}[t]{\linewidth}
				\begin{tcolorbox}[enhanced, colback=emerald!5, colframe=emerald!60!black, title=\faRobot \ \textsc{Qwen3-VL-8B}, fonttitle=\small\bfseries, attach boxed title to top left={yshift=-2mm, xshift=2mm}, boxed title style={colback=emerald!60!black}, boxrule=0.8pt, arc=4pt, fontupper=\scriptsize]
					\textbf{Reasoning:} The sign directly above the bicycle indicator (pedestrian and child) indicates a pedestrian zone. The bidirectional arrows on the auxiliary plate suggest this is a shared or pedestrian-priority area. The shaded path is likely restricted. \\
					\hrule \vspace{2pt}
					\textbf{Answer:} \textcolor{green!50!black}{\textbf{B. No, cycling is not allowed on this lane}}
				\end{tcolorbox}
			\end{minipage}

			 &

			\begin{minipage}[t]{\linewidth}
				\begin{tcolorbox}[enhanced, colback=cyan!5, colframe=cyan!60!black, title=\faRobot \ \textsc{InternVL3.5-8B}, fonttitle=\small\bfseries, attach boxed title to top left={yshift=-2mm, xshift=2mm}, boxed title style={colback=cyan!60!black}, boxrule=0.8pt, arc=4pt, fontupper=\scriptsize]
					\textbf{Reasoning:} The sign indicates that the lane is for both cyclists and pedestrians. Therefore, the cyclist is allowed to continue following the pink shaded lane forward as it is a shared space. \\

					\textbf{Answer:} \textcolor{red!80!black}{\textbf{A. Yes, continue riding forward on the shaded lane}}
				\end{tcolorbox}
			\end{minipage}
		\end{tabular}
	}

	\caption{\textbf{Failure case study for the Sign-Action Association (SAA) task.} Models fail to spatially perceive when multiple signs with different instructions are present next to each other, leading to confusion about which regulation applies to the shaded lane. Note that generated text is paraphrased for brevity.}
	\label{tab:casestudy13}

\end{table*}

\begin{table*}[!t]
	\centering

	\resizebox{\textwidth}{!}{%
		\begin{tabular}{@{} p{0.48\textwidth} p{0.52\textwidth} @{}}

			\begin{minipage}[c]{0.48\textwidth}
				\centering
                \includegraphics[width=\linewidth]{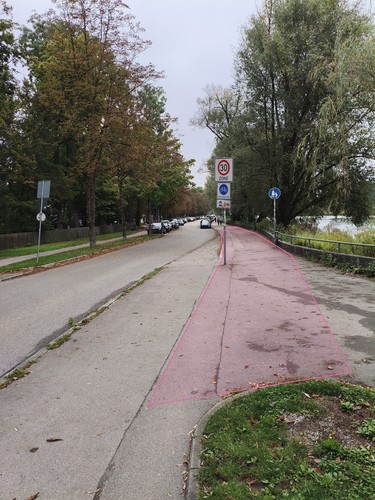}

			\end{minipage}
			 &
			\begin{minipage}[c]{0.52\textwidth}
				\begin{tcolorbox}[colback=blue!5, colframe=blue!50!black, title=\textbf{Question}, fonttitle=\small\bfseries, boxrule=0.8pt, arc=4pt, left=4pt, right=4pt, top=2pt, bottom=2pt, fontupper=\scriptsize]
					The image shows a forward-facing view from a cyclist’s perspective. Based on the shaded path (in pink) shown ahead, should the cyclist proceed forward on this lane? The shading is externally drawn on the image as an overlay.
				\end{tcolorbox}

				\begin{tcolorbox}[colback=gray!5, colframe=gray!40, title=\textbf{Choices}, fonttitle=\scriptsize\bfseries, coltitle=black, boxrule=0.5pt, arc=2pt, left=4pt, right=4pt, top=2pt, bottom=2pt, fontupper=\scriptsize]
					A. Yes, continue riding forward on the shaded lane \\
					B. No, cycling is not allowed on this lane \\
					\hrule \vspace{2pt}
					\textbf{\textcolor{green!50!black}{ GT:}} \textbf{B} — No, cycling is not allowed on this lane
				\end{tcolorbox}
			\end{minipage} \\
		\end{tabular}
	}

	\vspace{0.5em}
	\resizebox{\textwidth}{!}{%
		\begin{tabular}{@{} p{0.5\textwidth} p{0.5\textwidth} @{}}

			\begin{minipage}[t]{\linewidth}
				\begin{tcolorbox}[enhanced, colback=orange!5, colframe=orange!60!black, title=\faRobot \ \textsc{DriveLMM-O1 (8B)}, fonttitle=\small\bfseries, attach boxed title to top left={yshift=-2mm, xshift=2mm}, boxed title style={colback=orange!60!black}, boxrule=0.8pt, arc=4pt, fontupper=\scriptsize]
					\textbf{Reasoning:} The image shows a lane specifically designated for bicycles, indicated by the symbol on the sign above the lane. The pink mark suggests it is earmarked for cycling purposes. The cyclist's forward view implies the lane ahead is available. \\

					\textbf{Answer:} \textcolor{red!80!black}{\textbf{A. Yes, continue riding forward on the shaded lane}}
				\end{tcolorbox}
			\end{minipage}
			 &
			\begin{minipage}[t]{\linewidth}
				\begin{tcolorbox}[enhanced, colback=purple!5, colframe=purple!60!black, title=\faBrain \ \textsc{Cosmos-R2 (8B)}, fonttitle=\small\bfseries, attach boxed title to top left={yshift=-2mm, xshift=2mm}, boxed title style={colback=purple!60!black}, boxrule=0.8pt, arc=4pt, fontupper=\scriptsize]
					\textbf{Reasoning:} A blue sign with a white bicycle symbol indicates a bike lane. The presence of a "30 ZONE" sign suggests restricted car use. Since the shaded path is separated by a railing and labeled for cyclists, the cyclist should continue using it. \\

					\textbf{Answer:} \textcolor{red!80!black}{\textbf{A}}
				\end{tcolorbox}
			\end{minipage}
		\end{tabular}
	}

	\vspace{0.5em}
	\resizebox{\textwidth}{!}{%
		\begin{tabular}{@{} p{0.5\textwidth} p{0.5\textwidth} @{}}

			\begin{minipage}[t]{\linewidth}
				\begin{tcolorbox}[enhanced, colback=emerald!5, colframe=emerald!60!black, title=\faRobot \ \textsc{ReCogDrive-VLM}, fonttitle=\small\bfseries, attach boxed title to top left={yshift=-2mm, xshift=2mm}, boxed title style={colback=emerald!60!black}, boxrule=0.8pt, arc=4pt, fontupper=\scriptsize]
					\textbf{Reasoning:} The pink shade indicates a designated cycling path specifically for cyclists. This path is separate from the main road, ensuring a safe route. Therefore, the cyclist should proceed as it is clearly intended for cycling activities. \\

					\textbf{Answer:} \textcolor{red!80!black}{\textbf{A. Yes, continue riding forward on the shaded lane.}}
				\end{tcolorbox}
			\end{minipage}

			 &

			\begin{minipage}[t]{\linewidth}
				\begin{tcolorbox}[enhanced, colback=red!5, colframe=red!60!black, title=\faCube \ \textsc{SpatialThinker-7B}, fonttitle=\small\bfseries, attach boxed title to top left={yshift=-2mm, xshift=2mm}, boxed title style={colback=red!60!black}, boxrule=0.8pt, arc=4pt, fontupper=\scriptsize]
					\textbf{Reasoning:} The sign indicates a "Fahrradstraße" (bike lane) and a "30 ZONE" limit. The shaded pink path appears to be a designated bike lane, and the cyclist should follow it for safety and adherence to rules. \\

					\textbf{Answer:} \textcolor{red!80!black}{\textbf{A. Yes, continue riding forward on the shaded lane}}
				\end{tcolorbox}
			\end{minipage}
		\end{tabular}
	}

	\caption{\textbf{Failure case study for the Sign-Action Association (SAA) task.}		In this models fails to focus on the pedestrian lane sign on the right side. Note that generated text is paraphrased for brevity.}
	\label{tab:casestudy14}

\end{table*}

\begin{table*}[!t]
	\centering

	\resizebox{\textwidth}{!}{%
		\begin{tabular}{@{} p{0.48\textwidth} p{0.52\textwidth} @{}}

			\begin{minipage}[c]{0.48\textwidth}
				\centering
				\includegraphics[width=\linewidth]{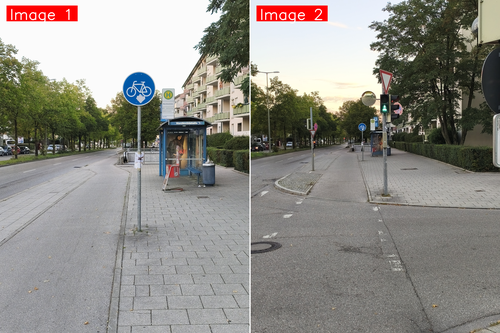}

			\end{minipage}
			 &
			\begin{minipage}[c]{0.52\textwidth}
				\begin{tcolorbox}[colback=blue!5, colframe=blue!50!black, title=\textbf{Question}, fonttitle=\small\bfseries, boxrule=0.8pt, arc=4pt, left=4pt, right=4pt, top=2pt, bottom=2pt, fontupper=\scriptsize]
					As the cyclist progressed along the road, which of these two views corresponds to the earlier moment?
				\end{tcolorbox}

				\begin{tcolorbox}[colback=gray!5, colframe=gray!40, title=\textbf{Choices}, fonttitle=\scriptsize\bfseries, coltitle=black, boxrule=0.5pt, arc=2pt, left=4pt, right=4pt, top=2pt, bottom=2pt, fontupper=\scriptsize]
					A. Image 1 \\
					B. Image 2 \\
					\hrule \vspace{2pt}
					\textbf{\textcolor{green!50!black}{ GT:}} \textbf{B} — Image 2 corresponds to the earlier moment.
				\end{tcolorbox}
			\end{minipage} \\
		\end{tabular}
	}

	\vspace{0.5em}
	\resizebox{\textwidth}{!}{%
		\begin{tabular}{@{} p{0.5\textwidth} p{0.5\textwidth} @{}}

			\begin{minipage}[t]{\linewidth}
				\begin{tcolorbox}[enhanced, colback=orange!5, colframe=orange!60!black, title=\faRobot \ \textsc{DriveLMM-O1 (8B)}, fonttitle=\small\bfseries, attach boxed title to top left={yshift=-2mm, xshift=2mm}, boxed title style={colback=orange!60!black}, boxrule=0.8pt, arc=4pt, fontupper=\scriptsize]
					\textbf{Reasoning:} Image 1 shows a clear route with fewer obstructions. Image 2 shows a more complex intersection with traffic signals, suggesting a more "developed" or later progression of the cyclist through the environment. \\

					\textbf{Answer:} \textcolor{red!80!black}{\textbf{A. Image 1}}
				\end{tcolorbox}
			\end{minipage}
			 &
			\begin{minipage}[t]{\linewidth}
				\begin{tcolorbox}[enhanced, colback=cyan!5, colframe=cyan!60!black, title=\faRobot \ \textsc{InternVL3.5-8B}, fonttitle=\small\bfseries, attach boxed title to top left={yshift=-2mm, xshift=2mm}, boxed title style={colback=cyan!60!black}, boxrule=0.8pt, arc=4pt, fontupper=\scriptsize]
					\textbf{Reasoning:} The blue bicycle sign is clearly visible in Image 1. Typically, such instructional signs are encountered earlier along a road as a cyclist progresses. Image 2 appears to be further along after passing this marker. \\

					\textbf{Answer:} \textcolor{red!80!black}{\textbf{A. Image 1}}
				\end{tcolorbox}
			\end{minipage}
		\end{tabular}
	}

	\vspace{0.5em}
	\resizebox{\textwidth}{!}{%
		\begin{tabular}{@{} p{0.5\textwidth} p{0.5\textwidth} @{}}

			\begin{minipage}[t]{\linewidth}
				\begin{tcolorbox}[enhanced, colback=purple!5, colframe=purple!60!black, title=\faBrain \ \textsc{Cosmos-R2 (8B)}, fonttitle=\small\bfseries, attach boxed title to top left={yshift=-2mm, xshift=2mm}, boxed title style={colback=purple!60!black}, boxrule=0.8pt, arc=4pt, fontupper=\scriptsize]
					\textbf{Reasoning:} Bus stops are usually encountered before major intersections. The person waiting at the shelter in Image 1 suggests an earlier point in the journey before the cyclist reaches the traffic control elements seen in Image 2. \\

					\textbf{Answer:} \textcolor{red!80!black}{\textbf{A}}
				\end{tcolorbox}
			\end{minipage}

			 &

			\begin{minipage}[t]{\linewidth}
				\begin{tcolorbox}[enhanced, colback=emerald!5, colframe=emerald!60!black, title=\faRobot \ \textsc{Ovis2.5-9B}, fonttitle=\small\bfseries, attach boxed title to top left={yshift=-2mm, xshift=2mm}, boxed title style={colback=emerald!60!black}, boxrule=0.8pt, arc=4pt, fontupper=\scriptsize]
					\textbf{Reasoning:} Prominent signage like the bicycle lane indicator in Image 1 marks an approaching zone. In Image 2, the cyclist has progressed to the intersection, approaching the traffic lights, which represents a later stage of the trip. \\

					\textbf{Answer:} \textcolor{red!80!black}{\textbf{A. Image 1}}
				\end{tcolorbox}
			\end{minipage}
		\end{tabular}
	}

	\vspace{0.5em}
	\resizebox{\textwidth}{!}{%
		\begin{tabular}{@{} p{0.5\textwidth} p{0.5\textwidth} @{}}

			\begin{minipage}[t]{\linewidth}
				\begin{tcolorbox}[enhanced, colback=blue!5, colframe=blue!60!black, title=\faRobot \ \textsc{Qwen3-VL-8B}, fonttitle=\small\bfseries, attach boxed title to top left={yshift=-2mm, xshift=2mm}, boxed title style={colback=blue!60!black}, boxrule=0.8pt, arc=4pt, fontupper=\scriptsize]
					\textbf{Reasoning:} Image 1 provides a direct, close-up view of the bus stop, suggesting the cyclist is just arriving at this point. Image 2 shows a wider perspective with more road ahead, implying the cyclist is further along. \\

					\textbf{Answer:} \textcolor{red!80!black}{\textbf{A. Image 1}}
				\end{tcolorbox}
			\end{minipage}

			 &

			\begin{minipage}[t]{\linewidth}
				\begin{tcolorbox}[enhanced, colback=gray!5, colframe=black, title=\faRobot \ \textsc{VST-7B-RL}, fonttitle=\small\bfseries, attach boxed title to top left={yshift=-2mm, xshift=2mm}, boxed title style={colback=black}, boxrule=0.8pt, arc=4pt, fontupper=\scriptsize]
					\textbf{Reasoning:} Image 1 captures a more central/early section of the road characterized by the bus stop. Image 2 captures a later section with different signage and a more open road environment. \\

					\textbf{Answer:} \textcolor{red!80!black}{\textbf{A}}
				\end{tcolorbox}
			\end{minipage}
		\end{tabular}
	}
	\caption{\textbf{Failure case study for the Temporal Ordering (TO) task.} Models misinterpret spatial cues and fail to accurately sequence the images based on the cyclist's progression along the road.
    Note that generated text is paraphrased for brevity.
    }
	\label{tab:casestudy15}
\end{table*}

\begin{table*}[!t]
	\centering

	\resizebox{\textwidth}{!}{%
		\begin{tabular}{@{} p{0.48\textwidth} p{0.52\textwidth} @{}}

			\begin{minipage}[c]{0.48\textwidth}
				\centering
				\includegraphics[width=\linewidth]{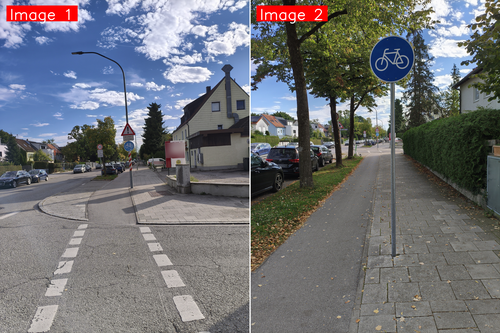}

			\end{minipage}
			 &
			\begin{minipage}[c]{0.52\textwidth}
				\begin{tcolorbox}[colback=blue!5, colframe=blue!50!black, title=\textbf{Question}, fonttitle=\small\bfseries, boxrule=0.8pt, arc=4pt, left=4pt, right=4pt, top=2pt, bottom=2pt, fontupper=\scriptsize]
					These two images were captured sequentially during the cyclist's forward travel. Which image was taken first?
				\end{tcolorbox}

				\begin{tcolorbox}[colback=gray!5, colframe=gray!40, title=\textbf{Choices}, fonttitle=\scriptsize\bfseries, coltitle=black, boxrule=0.5pt, arc=2pt, left=4pt, right=4pt, top=2pt, bottom=2pt, fontupper=\scriptsize]
					A. Image 1 \\
					B. Image 2 \\
					\hrule \vspace{2pt}
					\textbf{\textcolor{green!50!black}{ GT:}} \textbf{B} — Image 2 was taken first.
				\end{tcolorbox}
			\end{minipage} \\
		\end{tabular}
	}

	\vspace{0.5em}
	\resizebox{\textwidth}{!}{%
		\begin{tabular}{@{} p{0.5\textwidth} p{0.5\textwidth} @{}}

			\begin{minipage}[t]{\linewidth}
				\begin{tcolorbox}[enhanced, colback=orange!5, colframe=orange!60!black, title=\faRobot \ \textsc{DriveLMM-O1 (8B)}, fonttitle=\small\bfseries, attach boxed title to top left={yshift=-2mm, xshift=2mm}, boxed title style={colback=orange!60!black}, boxrule=0.8pt, arc=4pt, fontupper=\scriptsize]
					\textbf{Reasoning:} Image 1 shows a broad view of a curved road and intersection, suggesting the cyclist is approaching the path. Image 2 is a close-up of a sign, implying it was captured after moving past the broader scene. \\

					\textbf{Answer:} \textcolor{red!80!black}{\textbf{A. Image 1}}
				\end{tcolorbox}
			\end{minipage}
			 &
			\begin{minipage}[t]{\linewidth}
				\begin{tcolorbox}[enhanced, colback=purple!5, colframe=purple!60!black, title=\faBrain \ \textsc{Cosmos-R2 (8B)}, fonttitle=\small\bfseries, attach boxed title to top left={yshift=-2mm, xshift=2mm}, boxed title style={colback=purple!60!black}, boxrule=0.8pt, arc=4pt, fontupper=\scriptsize]
					\textbf{Reasoning:} The cyclist would first encounter the road intersection and speed signs (Image 1) before reaching the specific bike path indicated by the blue sign (Image 2) further along the journey. \\

					\textbf{Answer:} \textcolor{red!80!black}{\textbf{A}}
				\end{tcolorbox}
			\end{minipage}
		\end{tabular}
	}

	\vspace{0.5em}
	\resizebox{\textwidth}{!}{%
		\begin{tabular}{@{} p{0.5\textwidth} p{0.5\textwidth} @{}}

			\begin{minipage}[t]{\linewidth}
				\begin{tcolorbox}[enhanced, colback=emerald!5, colframe=emerald!60!black, title=\faRobot \ \textsc{Ovis2.5-9B}, fonttitle=\small\bfseries, attach boxed title to top left={yshift=-2mm, xshift=2mm}, boxed title style={colback=emerald!60!black}, boxrule=0.8pt, arc=4pt, fontupper=\scriptsize]
					\textbf{Reasoning:} Image 1 shows the cyclist at a street intersection, while Image 2 shows a sidewalk path. Since travel usually moves from main road intersections to residential paths, Image 1 was captured first. \\

					\textbf{Answer:} \textcolor{red!80!black}{\textbf{A. Image 1}}
				\end{tcolorbox}
			\end{minipage}

			 &

			\begin{minipage}[t]{\linewidth}
				\begin{tcolorbox}[enhanced, colback=blue!5, colframe=blue!60!black, title=\faRobot \ \textsc{Qwen3-VL-8B}, fonttitle=\small\bfseries, attach boxed title to top left={yshift=-2mm, xshift=2mm}, boxed title style={colback=blue!60!black}, boxrule=0.8pt, arc=4pt, fontupper=\scriptsize]
					\textbf{Reasoning:} Image 1 shows the cyclist approaching an intersection with road damage signs. Image 2 shows a straight path past that intersection. Given forward travel, the intersection view occurs first. \\

					\textbf{Answer:} \textcolor{red!80!black}{\textbf{A. Image 1}}
				\end{tcolorbox}
			\end{minipage}
		\end{tabular}
	}

	\vspace{0.5em}
	\resizebox{\textwidth}{!}{%
		\begin{tabular}{@{} p{0.5\textwidth} p{0.5\textwidth} @{}}

			\begin{minipage}[t]{\linewidth}
				\begin{tcolorbox}[enhanced, colback=red!5, colframe=red!60!black, title=\faCube \ \textsc{SpatialThinker-7B}, fonttitle=\small\bfseries, attach boxed title to top left={yshift=-2mm, xshift=2mm}, boxed title style={colback=red!60!black}, boxrule=0.8pt, arc=4pt, fontupper=\scriptsize]
					\textbf{Reasoning:} Image 1 shows the street from a distance, while Image 2 shows a close-up of a sign on the sidewalk. Forward travel implies getting closer to the sign, so the distant view is first. \\

					\textbf{Answer:} \textcolor{red!80!black}{\textbf{A (A) Image 1}}
				\end{tcolorbox}
			\end{minipage}

			 &

			\begin{minipage}[t]{\linewidth}
				\begin{tcolorbox}[enhanced, colback=gray!5, colframe=black, title=\faRobot \ \textsc{VST-7B-RL}, fonttitle=\small\bfseries, attach boxed title to top left={yshift=-2mm, xshift=2mm}, boxed title style={colback=black}, boxrule=0.8pt, arc=4pt, fontupper=\scriptsize]
					\textbf{Reasoning:} The road with vehicles and buildings (Image 1) represents the initial visual for a forward-moving cyclist before they transition to the specific sidewalk area (Image 2). \\

					\textbf{Answer:} \textcolor{red!80!black}{\textbf{A}}
				\end{tcolorbox}
			\end{minipage}
		\end{tabular}
	}
	\caption{\textbf{Failure case study for the Temporal Ordering (TO) task.} The models consistently misinterpreted the spatial progression of the cyclist, incorrectly identifying a more distant, complex intersection as the "earlier" moment and the immediate path sign as the "later" moment.
    Note that generated text is paraphrased for brevity.}
	\label{tab:casestudy16}

\end{table*}`'

\begin{table*}[!t]
	\centering

	\resizebox{\textwidth}{!}{%
		\begin{tabular}{@{} p{0.48\textwidth} p{0.52\textwidth} @{}}

			\begin{minipage}[c]{0.48\textwidth}
				\centering
				\includegraphics[width=\linewidth]{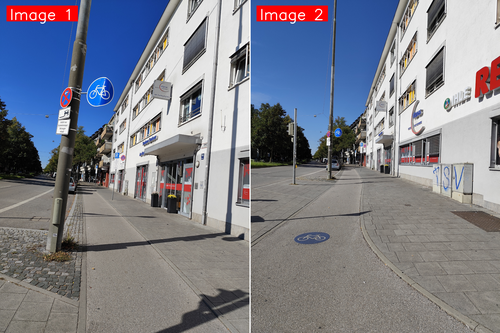}

			\end{minipage}
			 &
			\begin{minipage}[c]{0.52\textwidth}
				\begin{tcolorbox}[colback=blue!5, colframe=blue!50!black, title=\textbf{Question}, fonttitle=\small\bfseries, boxrule=0.8pt, arc=4pt, left=4pt, right=4pt, top=2pt, bottom=2pt, fontupper=\scriptsize]
					During continuous forward motion, the cyclist took these two photos. Which one was captured first in time?
				\end{tcolorbox}

				\begin{tcolorbox}[colback=gray!5, colframe=gray!40, title=\textbf{Choices}, fonttitle=\scriptsize\bfseries, coltitle=black, boxrule=0.5pt, arc=2pt, left=4pt, right=4pt, top=2pt, bottom=2pt, fontupper=\scriptsize]
					A. Image 1 \\
					B. Image 2 \\
					\hrule \vspace{2pt}
					\textbf{\textcolor{green!50!black}{ GT:}} \textbf{B} — Image 2 was captured first.
				\end{tcolorbox}
			\end{minipage} \\
		\end{tabular}
	}

	\vspace{0.5em}
	\resizebox{\textwidth}{!}{%
		\begin{tabular}{@{} p{0.5\textwidth} p{0.5\textwidth} @{}}

			\begin{minipage}[t]{\linewidth}
				\begin{tcolorbox}[enhanced, colback=emerald!5, colframe=emerald!60!black, title=\faRobot \ \textsc{DriveLMM-O1 (8B)}, fonttitle=\small\bfseries, attach boxed title to top left={yshift=-2mm, xshift=2mm}, boxed title style={colback=emerald!60!black}, boxrule=0.8pt, arc=4pt, fontupper=\scriptsize]
					\textbf{Reasoning:} Image 2 depicts the forward-facing road marking that the cyclist travels over, representing the path ahead. Image 1 captures specific side details and navigation signs near the curb. Given the forward motion, the road-path view in Image 2 occurs first. \\
					\hrule \vspace{2pt}
					\textbf{Answer:} \textcolor{green!50!black}{\textbf{B. Image 2}}
				\end{tcolorbox}
			\end{minipage}
			 &
			\begin{minipage}[t]{\linewidth}
				\begin{tcolorbox}[enhanced, colback=orange!5, colframe=orange!60!black, title=\faRobot \ \textsc{InternVL3.5-8B}, fonttitle=\small\bfseries, attach boxed title to top left={yshift=-2mm, xshift=2mm}, boxed title style={colback=orange!60!black}, boxrule=0.8pt, arc=4pt, fontupper=\scriptsize]
					\textbf{Reasoning:} Visual cues such as shadows and the positions of objects suggest that Image 1 was captured when the cyclist was closer to the buildings. Image 2 shows them further away, indicating it must have been taken later in the sequence. \\

					\textbf{Answer:} \textcolor{red!80!black}{\textbf{A. Image 1}}
				\end{tcolorbox}
			\end{minipage}
		\end{tabular}
	}

	\vspace{0.5em}
	\resizebox{\textwidth}{!}{%
		\begin{tabular}{@{} p{0.5\textwidth} p{0.5\textwidth} @{}}

			\begin{minipage}[t]{\linewidth}
				\begin{tcolorbox}[enhanced, colback=purple!5, colframe=purple!60!black, title=\faBrain \ \textsc{Cosmos-R2 (8B)}, fonttitle=\small\bfseries, attach boxed title to top left={yshift=-2mm, xshift=2mm}, boxed title style={colback=purple!60!black}, boxrule=0.8pt, arc=4pt, fontupper=\scriptsize]
					\textbf{Reasoning:} Image 1 captures the initial set of signs and a parked car. As the cyclist progresses, new elements like graffiti and broader building views appear in Image 2, suggesting the cyclist has moved past the initial parked car location. \\

					\textbf{Answer:} \textcolor{red!80!black}{\textbf{A}}
				\end{tcolorbox}
			\end{minipage}

			 &

			\begin{minipage}[t]{\linewidth}
				\begin{tcolorbox}[enhanced, colback=red!5, colframe=red!60!black, title=\faRobot \ \textsc{Ovis2.5-9B}, fonttitle=\small\bfseries, attach boxed title to top left={yshift=-2mm, xshift=2mm}, boxed title style={colback=red!60!black}, boxrule=0.8pt, arc=4pt, fontupper=\scriptsize]
					\textbf{Reasoning:} Image 1 shows a closer view of the building and signs, while Image 2 shows a more distant perspective. Since the cyclist is moving forward, the photo taken earlier would be the one closer to the initial landmarks. \\

					\textbf{Answer:} \textcolor{red!80!black}{\textbf{A. Image 1}}
				\end{tcolorbox}
			\end{minipage}
		\end{tabular}
	}

	\vspace{0.5em}
	\resizebox{\textwidth}{!}{%
		\begin{tabular}{@{} p{0.5\textwidth} p{0.5\textwidth} @{}}

			\begin{minipage}[t]{\linewidth}
				\begin{tcolorbox}[enhanced, colback=blue!5, colframe=blue!60!black, title=\faRobot \ \textsc{Qwen3-VL-8B}, fonttitle=\small\bfseries, attach boxed title to top left={yshift=-2mm, xshift=2mm}, boxed title style={colback=blue!60!black}, boxrule=0.8pt, arc=4pt, fontupper=\scriptsize]
					\textbf{Reasoning:} Image 1 shows the cyclist in closer proximity to the signpost. Image 2 shows the cyclist further down the street with a bike symbol on the road ahead. The earlier photo would naturally be closer to the starting reference point. \\

					\textbf{Answer:} \textcolor{red!80!black}{\textbf{A. Image 1}}
				\end{tcolorbox}
			\end{minipage}

			 &

			\begin{minipage}[t]{\linewidth}
				\begin{tcolorbox}[enhanced, colback=gray!5, colframe=red!60!black, title=\faCube \ \textsc{SpatialThinker-7B}, fonttitle=\small\bfseries, attach boxed title to top left={yshift=-2mm, xshift=2mm}, boxed title style={colback=red!60!black}, boxrule=0.8pt, arc=4pt, fontupper=\scriptsize]
					\textbf{Reasoning:} In Image 1, the cyclist is closer to the building and immediate surroundings. In Image 2, the view is more open and shows a wider street. Therefore, the closer view in Image 1 must have been captured first. \\

					\textbf{Answer:} \textcolor{red!80!black}{\textbf{A. Image 1}}
				\end{tcolorbox}
			\end{minipage}
		\end{tabular}
	}

	\caption{\textbf{Failure case study for the Temporal Ordering (TO) task.} Here, most models assume that being near a building or signpost represents the start of a sequence, rather than correctly identifying the trafic elements present in the scene to base its reasoning.
    Note that generated text is paraphrased for brevity.}
	\label{tab:casestudy17}

\end{table*}

\end{document}